%% file: main.tex
\documentclass{article}

\PassOptionsToPackage{numbers}{natbib}

\usepackage[final]{neurips_2021}

\usepackage[utf8]{inputenc} 
\usepackage[T1]{fontenc}    
\usepackage{hyperref}       
\usepackage{url}            
\usepackage{booktabs}       
\usepackage{amsfonts}       
\usepackage{nicefrac}       
\usepackage{microtype}      
\usepackage[table]{xcolor}         

\usepackage{graphicx}
\usepackage{multirow}
\usepackage{subcaption}
 \usepackage{changes}
\usepackage{amsmath}
\usepackage{amssymb}
\usepackage{nicefrac}
\usepackage{bbm}
\usepackage{wrapfig}

\newcommand{\methodname}{AugSelf}

\hypersetup{colorlinks,citecolor={blue}}

\definecolor{Gray}{gray}{0.9}
\newcommand{\facc}[2]{{#1}{\scriptsize±{#2}}}
\newcommand{\bfacc}[2]{\textbf{{#1}{\scriptsize±{#2}}}}
\definechangesauthor[color=purple, name=kimin]{KM}
\definechangesauthor[color=teal, name=kibok]{KB}
\definechangesauthor[color=magenta, name=hankook]{HK}

\title{Improving Transferability of Representations \\ via Augmentation-Aware Self-Supervision}

\author{%
Hankook Lee$^1$\quad Kibok Lee$^{23}$\thanks{Work done while at University of Michigan.}\quad Kimin Lee$^4$\quad Honglak Lee$^{25}$\quad Jinwoo Shin$^1$ \\
$^1$Korea Advanced Institute of Science and Technology (KAIST) \\
$^2$University of Michigan \\
$^3$Amazon Web Services \\
$^4$University of California, Berkeley \\
$^5$LG AI Research
}

\begin{document}

\maketitle

\begin{abstract}
Recent unsupervised representation learning methods have shown to be effective in a range of vision tasks by learning representations invariant to data augmentations such as random cropping and color jittering. However, such invariance could be harmful to downstream tasks if they rely on the characteristics of the data augmentations, e.g., location- or color-sensitive. This is not an issue just for unsupervised learning; we found that this occurs even in supervised learning because it also learns to predict the same label for all augmented samples of an instance. To avoid such failures and obtain more generalizable representations, we suggest to optimize an auxiliary self-supervised loss, coined \emph{\methodname}, that learns the difference of augmentation parameters (e.g., cropping positions, color adjustment intensities) between two randomly augmented samples. Our intuition is that {\methodname} encourages to preserve augmentation-aware information in learned representations, which could be beneficial for their transferability. Furthermore, {\methodname} can easily be incorporated into recent state-of-the-art representation learning methods with a negligible additional training cost. Extensive experiments demonstrate that our simple idea consistently improves the transferability of representations learned by supervised and unsupervised methods in various transfer learning scenarios. The code is available at \url{https://github.com/hankook/AugSelf}.
\end{abstract}

\input{1_intro}

\input{2_background}
\input{3_method}

\input{4_experiment}

\input{5_related}

\section{Discussion and conclusion}
To improve the transferability of representations, we propose {\methodname}, an auxiliary augmentation-aware self-supervision method that encourages representations to contain augmentation-aware information that could be useful in downstream tasks. Our idea is to learn to predict the difference between augmentation parameters of two randomly augmented views. Through extensive experiments, we demonstrate the effectiveness of {\methodname} in various transfer learning scenarios. We believe that our work would guide many research directions for unsupervised representation learning and transfer learning.

\textbf{Limitations.}
Even though our method provides large gains when we apply it to popular data augmentations like random cropping, it might not be applicable to some specific data augmentation, where it is non-trivial to parameterize augmentation such as GAN-based one \cite{sandfort2019data}. Hence, an interesting future direction would be to develop an augmentation-agnostic way that does not require to explicitly design augmentation-specific self-supervision. Even for this direction, we think that our idea of \emph{learning the relation between two augmented samples} could be used, e.g., constructing contrastive pairs between the relations.

\textbf{Negative societal impacts.}
Self-supervised training typically requires a huge training cost (e.g., training MoCo v2 \cite{chen2020moco_v2} for 1000 epochs requires 11 days on 8 V100 GPUs), a large network capacity (e.g., GPT3 \cite{brown2020gpt3} requires 175 billion parameters), therefore it raises environmental issues such as carbon generation \cite{schwartz2019green}. Hence, efficient training methods \cite{wang2021solving} or distilling knowledge to a smaller network \cite{fang2021seed} would be required to ameliorate such environmental problems.

\section*{Acknowledgments and disclosure of funding}

This work was mainly supported by Samsung Electronics Co., Ltd (IO201211-08107-01) and partly supported by Institute of Information \& Communications Technology Planning \& Evaluation (IITP) grant funded by the Korea government (MSIT) (No.2019-0-00075, Artificial Intelligence Graduate School Program (KAIST)).

\bibliography{main}
\bibliographystyle{unsrtnat}


\newpage
\appendix
\input{0_appendix}

\end{document}

%% file: 1_intro.tex
\section{Introduction}\label{section:introduction}

Unsupervised representation learning has recently shown a remarkable success in various domains, e.g., computer vision \cite{he2020moco,chen2020simclr,han2020self_video}, natural language \cite{devlin2018bert,brown2020gpt3}, code \cite{feng2020codebert}, reinforcement learning \cite{Anand2019unsup_atari,stooke2020decoupling,Laskin20curl}, and graphs \cite{hassani2020contrastive_graph}. The representations pretrained with a large number of unlabeled data have achieved outstanding performance in various downstream tasks, by either training task-specific layers on top of the model while freezing it or fine-tuning the entire model.

In the vision domain, the recent state-of-the-art methods \citep{he2020moco,chen2020simclr,caron2020swav,grill2020byol,chen2020simsiam} learn representations to be invariant to a pre-defined set of augmentations. The choice of the augmentations plays a crucial role in representation learning \citep{chen2020simclr,chen2020moco_v2,tian2020good_view,xiao2021what}. A common choice is a combination of random cropping, horizontal flipping, color jittering, grayscaling, and Gaussian blurring. With this choice, learned representations are invariant to color and positional information in images; in other words, the representations lose such information.

On the contrary, there have also been attempts to learn representations by designing pretext tasks that keep such information in augmentations, e.g., predicting positional relations between two patches of an image \cite{doersch2015patch_location}, solving jigsaw puzzles \cite{noroozi2016jigsaw}, or predicting color information from a gray image \cite{zhang2016colorful}.
These results show the importance of augmentation-specific information for representation learning, and inspire us to explore the following research questions: 
\emph{when is learning invariance to a given set of augmentations harmful to representation learning?} and, \emph{how to prevent the loss in the recent unsupervised learning methods?}

\input{fig_concept}
\textbf{Contribution.} 
We first found that learning representations with an augmentation-invariant objective might hurt its performance in downstream tasks that rely on information related to the augmentations. For example, learning invariance against strong color augmentations forces the representations to contain less color information (see Figure \ref{figure:motivation:mi}). Hence, it degrades the performance of the representations in color-sensitive downstream tasks such as the Flowers classification task \cite{nilsback2008data_flowers102} (see Figure~\ref{figure:motivation:flower}).

To prevent this information loss and obtain more generalizable representations, we propose an auxiliary self-supervised loss, coined {\methodname}, that learns the difference of augmentation parameters between the two augmented samples (or views) as shown in Figure \ref{figure:concept}. For example, in the case of random cropping, {\methodname} learns to predict the difference of cropping positions of two randomly cropped views. 
We found that {\methodname} encourages the self-supervised representation learning methods, such as SimCLR \cite{chen2020simclr} and SimSiam \cite{chen2020simsiam}, to preserve augmentation-aware information (see Figure \ref{figure:motivation:mi}) that could be useful for downstream tasks. Furthermore, {\methodname} can easily be incorporated into the recent unsupervised representation learning methods \citep{he2020moco,chen2020simclr,grill2020byol,chen2020simsiam} with a negligible additional training cost, which is for training an auxiliary prediction head $\phi$ in Figure \ref{figure:concept}. 

Somewhat interestingly, we also found that optimizing the auxiliary loss, {\methodname}, can even improve the transferability of representations learned under the standard supervised representation learning scenarios \cite{sharif2014cnn}. This is because supervised learning also forces invariance, i.e., assigns the same label, for all augmented samples (of the same instance), and {\methodname} can help to keep augmentation-aware knowledge in the learned representations.

We demonstrate the effectiveness of {\methodname} under extensive transfer learning experiments: {\methodname} improves (a) two unsupervised representation learning methods, MoCo \cite{he2020moco} and SimSiam \cite{chen2020simsiam}, in 20 of 22 tested scenarios; (b) supervised pretraining in 9 of 11 (see Table \ref{table:transfer_linear}). Furthermore, we found that {\methodname} is also effective under few-shot learning setups (see Table \ref{table:transfer_fewshot}).

Remark that learning augmentation-invariant representations has been a common practice for both supervised and unsupervised representation learning frameworks, while the importance of augmentation-awareness is less emphasized. We hope that our work could inspire researchers to rethink the under-explored aspect and provide a new angle in representation learning.

%% file: fig_concept.tex
\begin{figure}[t]
\centering
\includegraphics[width=0.82\textwidth]{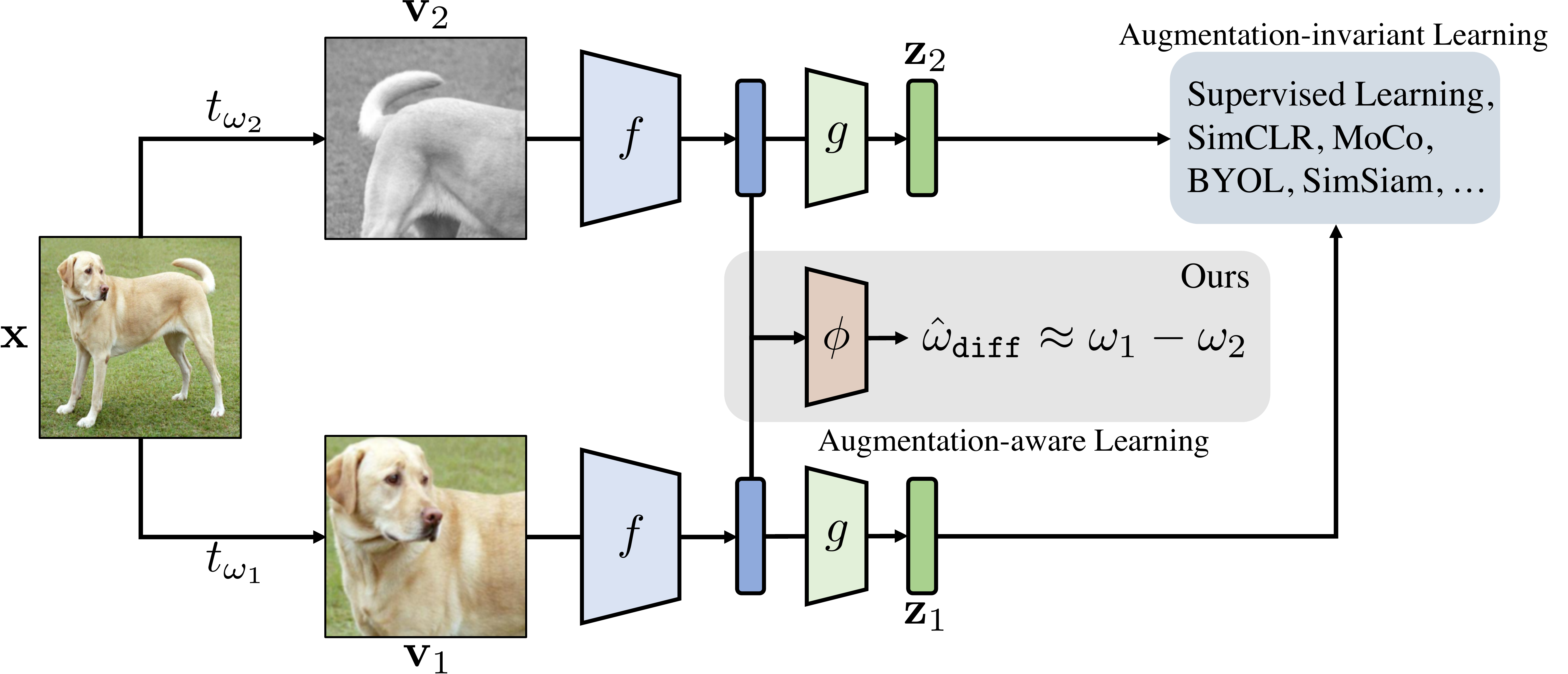}
\caption{Illustration of the proposed method, {\methodname}, that learns augmentation-aware information by predicting the difference between two augmentation parameters $\omega_1$ and $\omega_2$.
Here, $\mathbf{x}$ is an original image, $\mathbf{v}=t_\omega(\mathbf{x})$ is an augmented sample by an augmentation $t_\omega$, $f$ is a feature extractor such as ResNet \citep{he2016resnet}, and $g$ is a classifier for supervised learning or a projection MLP head for the recent unsupervised learning methods \citep{he2020moco,chen2020simclr,grill2020byol,chen2020simsiam}. 
}\label{figure:concept}
\vspace{-0.15in}
\end{figure}

%% file: 2_background.tex
\section{Preliminaries: Augmentation-invariant representation learning}\label{section:background}

\input{fig_mi}

In this section, we review the recent unsupervised representation learning methods \citep{he2020moco,chen2020simclr,caron2020swav,grill2020byol,chen2020simsiam} that learn representations by optimizing augmentation-invariant objectives. Formally, let $\mathbf{x}$ be an image, $t_\omega$ be an augmentation function parameterized by an augmentation parameter $\omega$, $\mathbf{v}=t_{\omega}(\mathbf{x})$ be the augmented sample (or view) of $\mathbf{x}$ by $t_\omega$, and $f$ be a CNN feature extractor, such as ResNet \citep{he2016resnet}. Generally speaking, the methods encourage the representations $f(\mathbf{v}_1)$ and $f(\mathbf{v}_2)$ to be invariant to the two randomly augmented views $\mathbf{v}_1=t_{\omega_1}(\mathbf{x})$ and $\mathbf{v}_2=t_{\omega_2}(\mathbf{x})$, i.e., $f(\mathbf{v}_1)\approx f(\mathbf{v}_2)$ for $\omega_1,\omega_2\sim\Omega$ where $\Omega$ is a pre-defined augmentation parameter distribution. We now describe the recent methods one by one briefly. For simplicity, we here omit the projection MLP head $g(\cdot)$ which is widely used in the methods (see Figure \ref{figure:concept}).

\textbf{Instance contrastive learning approaches} \citep{he2020moco,chen2020simclr,wu2018instance_disc} minimize the distance between an anchor $f(t_{\omega_1}(\mathbf{x}))$ and its positive sample $f(t_{\omega_2}(\mathbf{x}))$, while maximizing the distance between the anchor $f(t_{\omega_1}(\mathbf{x}))$ and its negative sample $f(t_{\omega_3}(\mathbf{x}^\prime))$. Since contrastive learning performance depends on the number of negative samples, a memory bank \cite{wu2018instance_disc}, a large batch \cite{chen2020simclr}, or a momentum network with a representation queue \cite{he2020moco} has been utilized.

\textbf{Clustering approaches} \cite{caron2020swav,caron2018deepcluster,asano2019sela} encourage two representations $f(t_{\omega_1}(\mathbf{x}))$ and $f(t_{\omega_2}(\mathbf{x}))$ to be assigned into the same cluster, in other words, the distance between them will be minimized.

\textbf{Negative-free methods} \cite{grill2020byol,chen2020simsiam}
learn to predict the representation $f(\mathbf{v}_1)$ of a view $\mathbf{v}_1=t_{\omega_1}(\mathbf{x})$ from another view $\mathbf{v}_2=t_{\omega_2}(\mathbf{x})$.
For example, SimSiam \citep{chen2020simsiam} minimizes $\lVert h(f(\mathbf{v}_2))-\mathtt{sg}(f(\mathbf{v}_1))\rVert_2^2$ where $h$ is
an MLP and $\mathtt{sg}$ is the stop-gradient operation. In these methods, if $h$ is optimal, then $h(f(\mathbf{v}_2))=\mathbb{E}_{\omega_1\sim\Omega}[f(\mathbf{v_1})]$; thus, the expectation of the objective can be rewritten as $\mathrm{Var}_{\omega\sim\Omega}(f(\mathbf{v}))$. Therefore, the methods can be considered as learning invariance with respect to the augmentations. 

\textbf{Supervised learning approaches} \cite{sharif2014cnn} also learn augmentation-invariant representations. Since they often maximize $\exp(\mathbf{c}^\top_y f(t(\mathbf{x})))/\sum_{y^\prime}\exp(\mathbf{c}^\top_{y^\prime} f(t(\mathbf{x})))$ where $\mathbf{c}_y$ is the prototype vector of the label $y$, $f(t(\mathbf{x}))$ is concentrated to $\mathbf{c}_y$, i.e., $\mathbf{c}_y\approx f(t_{\omega_1}(\mathbf{x}))\approx f(t_{\omega_2}(\mathbf{x}))$.

These approaches encourage representations $f(\mathbf{x})$ to contain shared (i.e., augmentation-invariant) information between $t_{\omega_1}(\mathbf{x})$ and $t_{\omega_2}(\mathbf{x})$ and discard other information \cite{tian2020good_view}.
For example, if $t_\omega$ changes color information, then to satisfy $f(t_{\omega_1}(\mathbf{x}))=f(t_{\omega_2}(\mathbf{x}))$ for any $\omega_1, \omega_2 \sim \Omega$, $f(\mathbf{x})$ will be learned to contain no (or less) color information.
To verify this, we pretrain ResNet-18 \cite{he2016resnet} on STL10 \cite{krizhevsky2009cifar} using SimSiam \cite{chen2020simsiam}  with varying the strength $s$ of the color jittering augmentation. To measure the mutual information between representations and color information, we use the InfoNCE loss \cite{oord2018cpc}. We here simply encode color information as RGB color histograms of an image. As shown in Figure \ref{figure:motivation:mi}, using stronger color augmentations leads to color-relevant information loss. In classification on Flowers \cite{nilsback2008data_flowers102} and Food \cite{bossard14data_food101}, which is color-sensitive, the learned representations containing less color information result in lower performance as shown in Figure \ref{figure:motivation:flower} and \ref{figure:motivation:food}, respectively. This observation emphasizes the importance of learning augmentation-aware information in transfer learning scenarios.

%% file: fig_mi.tex
\begin{figure}[t]
\centering
\begin{subfigure}[b]{0.32\textwidth}
\centering
\includegraphics[width=\textwidth]{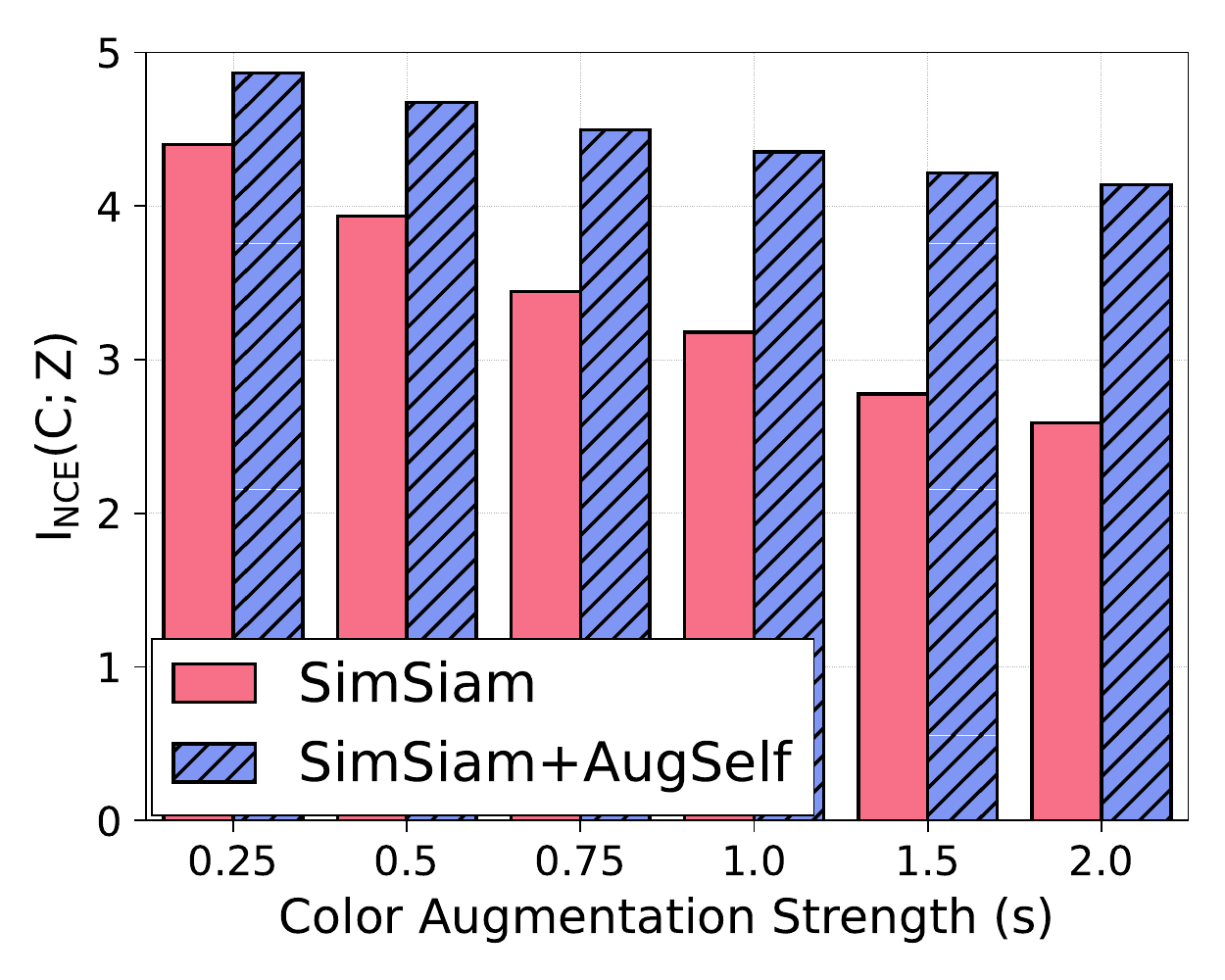}
\caption{Mutual information}\label{figure:motivation:mi}
\end{subfigure}
\begin{subfigure}[b]{0.32\textwidth}
\centering
\includegraphics[width=\textwidth]{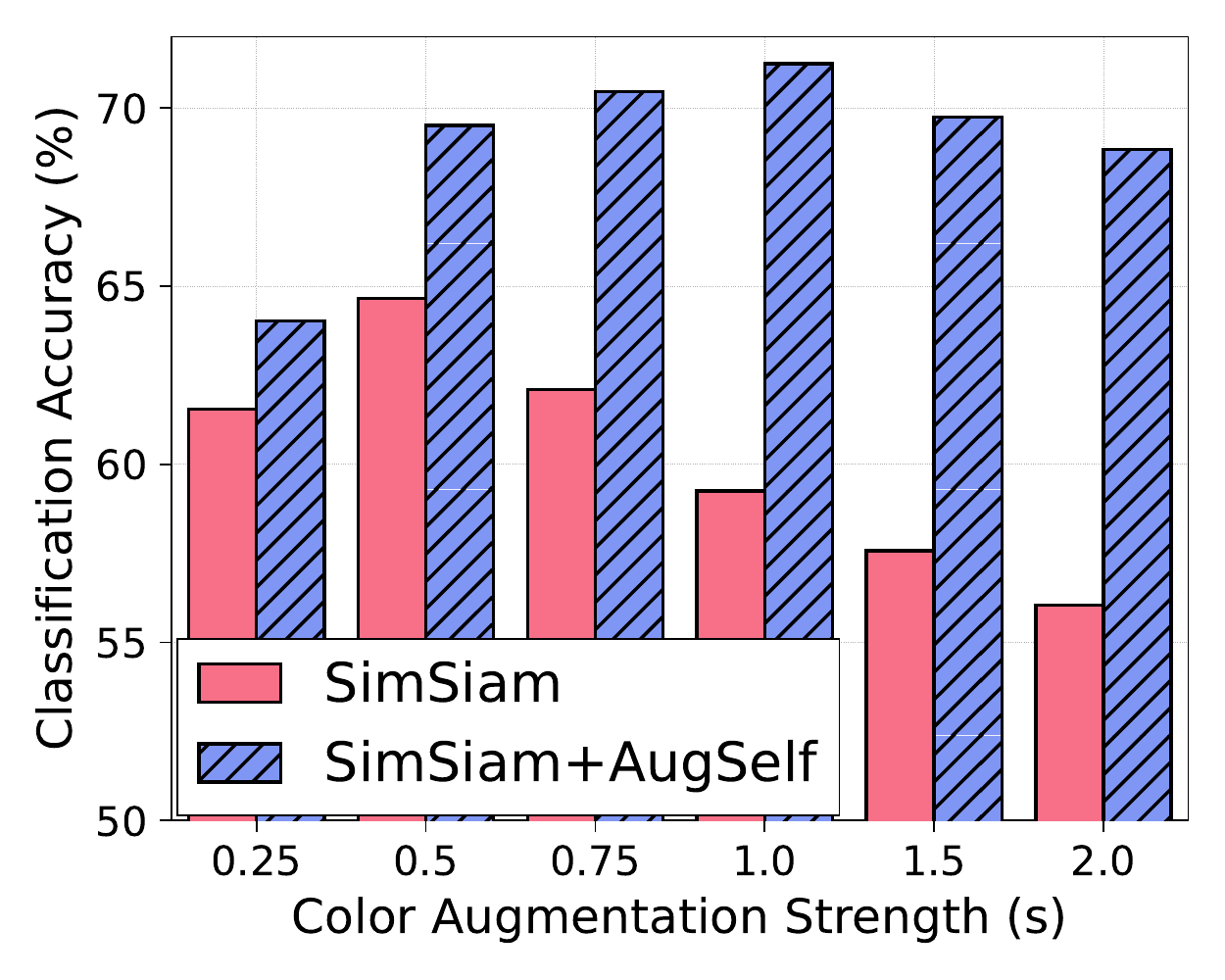}
\caption{STL10$\rightarrow$Flowers}\label{figure:motivation:flower}
\end{subfigure}
\begin{subfigure}[b]{0.32\textwidth}
\centering
\includegraphics[width=\textwidth]{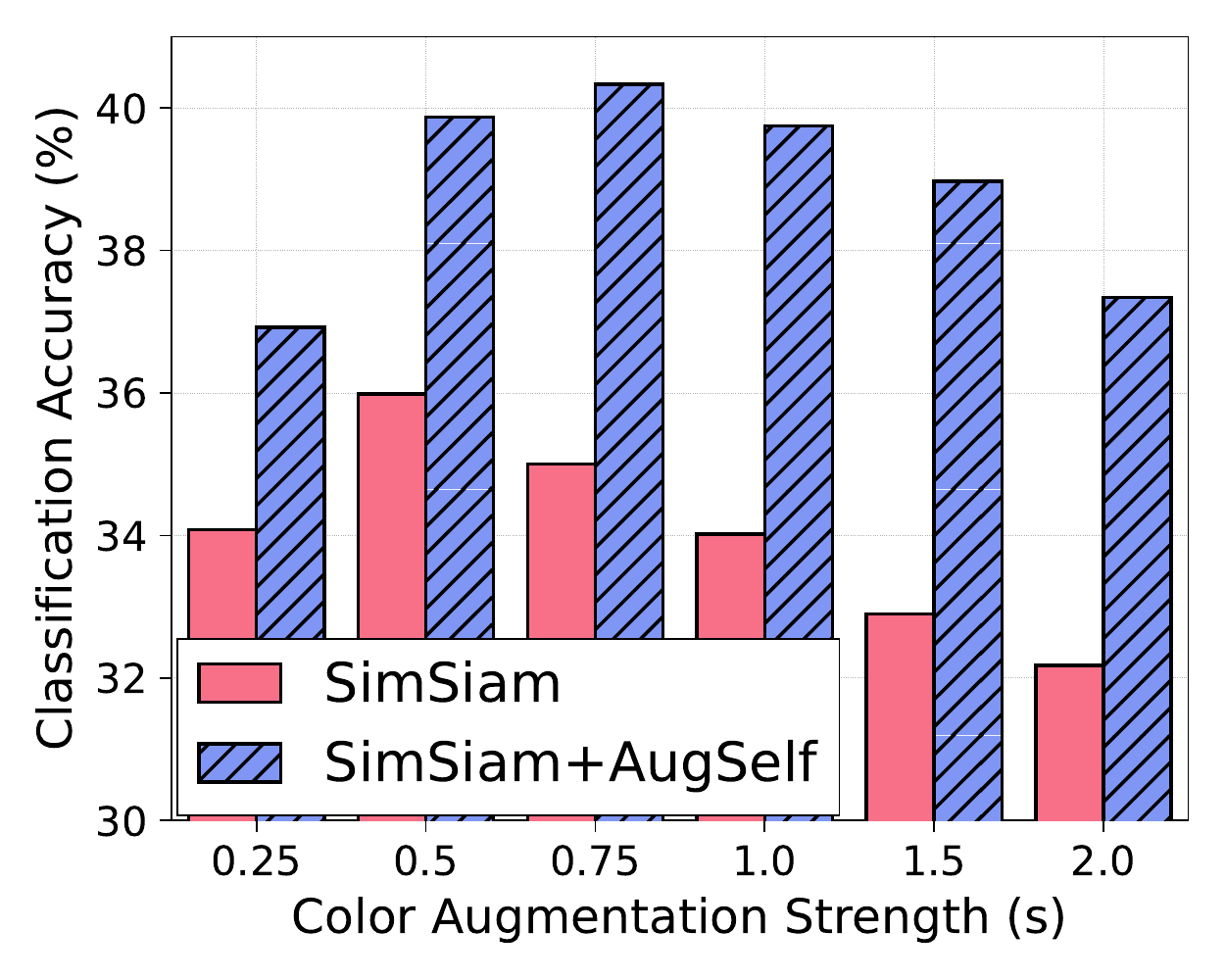}
\caption{STL10$\rightarrow$Food}\label{figure:motivation:food}
\end{subfigure}
\caption{(a) Changes of mutual information, i.e., $I_{\tt NCE}(C;\mathbf{z})$, between color information $C(\mathbf{x})$ and the representation $\mathbf{z}=f(\mathbf{x})$ pretrained on STL10 \cite{coates2011stl10} with varying the color jittering strength $s$. The pretrained representations are evaluated in color-sensitive benchmarks, (b) Flowers \cite{nilsback2008data_flowers102} and (c) Food \cite{bossard14data_food101}, by the linear evaluation protocol \cite{kornblith2019better}.}
\label{figure:motivation}
\vspace{-0.1in}
\end{figure}

%
%
%
%
%
%

%% file: 3_method.tex
\section{Auxiliary augmentation-aware self-supervision}\label{section:method}

\input{fig_augs}

In this section, we introduce \emph{auxiliary augmentation-aware self-supervision}, coined \emph{\methodname}, which encourages to preserve augmentation-aware information for generalizable representation learning. To be specific, 
we add an auxiliary self-supervision loss, which learns to predict the difference between augmentation parameters of two randomly augmented views, into existing augmentation-invariant representation learning methods \cite{he2020moco,chen2020simclr,caron2020swav,grill2020byol,chen2020simsiam}. We first describe a general form of our auxiliary loss, and then specific forms for various augmentations.
For conciseness, let $\theta$ be the collection of all parameters in the model.

Since an augmentation function $t_\omega$ is typically a composition of different types of augmentations, the augmentation parameter $\omega$ can be written as  $\omega=(\omega^\mathtt{aug})_{\mathtt{aug}\in\mathcal{A}}$
where $\mathcal{A}$ is the set of augmentations used in pretraining (e.g., $\mathcal{A}=\{\mathtt{crop},\mathtt{flip}$\}),
and $\omega^\mathtt{aug}$ is an augmentation-specific parameter (e.g., $\omega^\mathtt{crop}$ decides how to crop an image). Then, given two randomly augmented views $\mathbf{v}_1=t_{\omega_1}(\mathbf{x})$ and $\mathbf{v}_2=t_{\omega_2}(\mathbf{x})$, the {\methodname} objective is as follows:
\begin{align*}
\mathcal{L}_\mathtt{\methodname}(\mathbf{x},\omega_1,\omega_2;\theta)
=\sum\nolimits_{\mathtt{aug}\in\mathcal{A}_\mathtt{\methodname}}\mathcal{L}_\mathtt{aug}\big(\phi_\theta^{\mathtt{aug}}(f_\theta(\mathbf{v}_1),f_\theta(\mathbf{v}_2)),\omega^{\mathtt{aug}}_{\mathtt{diff}}\big),
\end{align*}
where $\mathcal{A}_\mathtt{\methodname}\subseteq\mathcal{A}$ is the set of augmentations for augmentation-aware learning, $\omega^\mathtt{aug}_\mathtt{diff}$ is the difference between two augmentation-specific parameters $\omega_1^\mathtt{aug}$ and $\omega_2^\mathtt{aug}$, $\mathcal{L}_\mathtt{aug}$ is an augmentation-specific loss, and $\phi_\theta^\mathtt{aug}$ is a 3-layer MLP for $\omega_\mathtt{diff}^\mathtt{aug}$ prediction. 
This design allows us to incorporate {\methodname} into the recent state-of-the-art unsupervised learning methods \citep{he2020moco,chen2020simclr,caron2020swav,grill2020byol,chen2020simsiam}  with a negligible additional training cost. For example, the objective of SimSiam \citep{chen2020simsiam} with {\methodname} can be written as
$\mathcal{L}_\mathtt{total}(\mathbf{x},\omega_1,\omega_2;\theta)
=\mathcal{L}_\mathtt{SimSiam}(\mathbf{x},\omega_1,\omega_2;\theta)+\lambda\cdot\mathcal{L}_\mathtt{\methodname}(\mathbf{x},\omega_1,\omega_2;\theta),$
where $\lambda$ is a hyperparameter for balancing losses. Remark that the total objective $\mathcal{L}_\mathtt{total}$ encourages the shared representation $f(\mathbf{x})$ to learn both augmentation-invariant and augmentation-aware features. Hence, the learned representation $f(\mathbf{x})$ also can be useful in various downstream (e.g., augmentation-sensitive) tasks.\footnote{We observe that our augmentation-aware objective $\mathcal{L}_\mathtt{\methodname}$ does not interfere with learning the augmentation-invariant objective, e.g., $\mathcal{L}_\mathtt{SimSiam}$. This allows $f(\mathbf{x})$ to learn augmentation-aware information with a negligible loss of augmentation-invariant information. A detailed discussion is provided in Appendix \ref{appendix:tradeoff}.}

In this paper, we mainly focus on the commonly-used augmentations in the recent unsupervised representation learning methods~\citep{he2020moco,chen2020simclr,caron2020swav,grill2020byol,chen2020simsiam}: random cropping, random horizontal flipping, color jittering, and Gaussian blurring; however, we remark that different types of 
augmentations can be incorporated into {\methodname} (see Section \ref{section:ablation}). In the following, we elaborate on the details of $\omega^\mathtt{aug}$ and $\mathcal{L}_\mathtt{aug}$ for each augmentation. The examples of $\omega^\mathtt{aug}$ are illustrated in Figure \ref{figure:augs}. 

\textbf{Random cropping.} The random cropping is the most popular augmentation in vision tasks. A cropping parameter $\omega^\mathtt{crop}$ contains the center position and cropping size. We
normalize the values by the height and width of the original image $\mathbf{x}$, i.e., $\omega^\mathtt{crop}\in[0,1]^4$. Then, we use $\ell_2$ loss for $\mathcal{L}_\mathtt{crop}$ and set $\omega_\mathtt{diff}^\mathtt{crop}=\omega^\mathtt{crop}_1-\omega^\mathtt{crop}_2$. 

\textbf{Random horizontal flipping.} A flipping parameter $\omega^\mathtt{flip}\in\{0,1\}$ indicates the image is horizontally flipped or not. Since it is discrete, we use the binary cross-entropy loss for $\mathcal{L}_\mathtt{flip}$ and set $\omega_\mathtt{diff}^\mathtt{flip}=\mathbbm{1}[{\omega^\mathtt{flip}_1=\omega^\mathtt{flip}_2}]$. 

\textbf{Color jittering.} The color jittering augmentation adjusts brightness, contrast, saturation, and hue of an input image in a random order. For each adjustment, its intensity is uniformly sampled from a pre-defined interval. We normalize all intensities into $[0,1]$, i.e., $\omega^\mathtt{color}\in[0,1]^4$. Similarly to cropping, we use $\ell_2$ loss for $\mathcal{L}_\mathtt{color}$ and set $\omega^\mathtt{color}_\mathtt{diff}=\omega^\mathtt{color}_1{-}\omega^\mathtt{color}_2$. 

\textbf{Gaussian blurring.} This blurring operation is widely used in unsupervised representation learning. The Gaussian filter is constructed by a single parameter, standard deviation $\sigma=\omega^\mathtt{blur}$. We also normalize the parameter into $[0,1]$. Then, we use $\ell_2$ loss for $\mathcal{L}_\mathtt{blur}$ and set $\omega^\mathtt{blur}_\mathtt{diff}=\omega^\mathtt{blur}_1{-}\omega^\mathtt{blur}_2$.

%% file: fig_augs.tex
\begin{figure*}[t]
\centering
\includegraphics[width=0.95\textwidth]{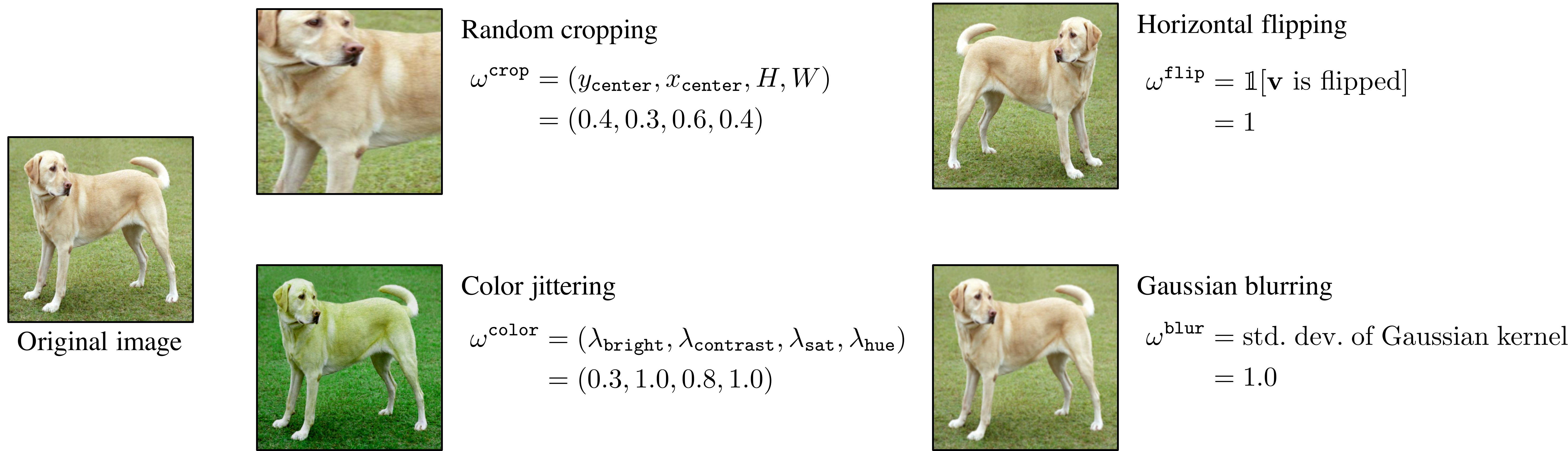}
\caption{Examples of the commonly-used augmentations and their parameters $\omega^\mathtt{aug}$.}\label{figure:augs}
\end{figure*}

%% file: 4_experiment.tex
\section{Experiments}\label{section:exp}

\textbf{Setup.} We pretrain the standard ResNet-18 \cite{he2016resnet} and
ResNet-50 on STL10 \cite{coates2011stl10} and ImageNet100\footnote{ImageNet100 is a 100-category subset of ImageNet \cite{russakovsky2015imagenet}. We use the same split following \citet{tian2019cmc}.} \cite{russakovsky2015imagenet,tian2019cmc}, respectively. We use two recent unsupervised representation learning methods as baselines for pretraining: a contrastive method, MoCo v2 \cite{he2020moco,chen2020moco_v2}, and a non-contrastive method, SimSiam \cite{chen2020simsiam}.
For STL10 and ImageNet100, we pretrain networks for $200$ and $500$ epochs with a batch size of $256$, respectively. 
For supervised pretraining, we pretrain ResNet-50 for $100$ epochs with a batch size of $128$ on ImageNet100.\footnote{We do not experiment supervised pretraining on STL10, as it has only 5k labeled training samples, which is not enough for pretraining a good representation.}
For augmentations, we use random cropping, flipping, color jittering, grayscaling, and Gaussian blurring following \citet{chen2020simsiam}. 
In this section, our {\methodname} predicts random cropping and color jittering parameters, i.e., $\mathcal{A}_\mathtt{\methodname}=\{\mathtt{crop},\mathtt{color}\}$, unless otherwise stated. We set $\lambda=1.0$ for STL10 and $\lambda=0.5$ for ImageNet100. The other details and the sensitivity analysis to the hyperparameter $\lambda$ are provided in Appendix \ref{sec:supp:pretrain} and \ref{appendix:lambda}, respectively. For ablation study (Section \ref{section:ablation}), we only use STL10-pretrained models.

\input{table_imagenet100_lineval}

\subsection{Main results}\label{section:imagenet100}

\input{table_imagenet100_fewshot}

\textbf{Linear evaluation in various downstream tasks.} 
We evaluate the pretrained networks in downstream classification tasks on 11 datasets: 
CIFAR10/100 \cite{krizhevsky2009cifar}, 
Food \cite{bossard14data_food101}, 
MIT67 \cite{quattoni2009mit67}, 
Pets \cite{parkhi2012pets}, 
Flowers \cite{nilsback2008data_flowers102}, 
Caltech101 \cite{fei2004calteck101},
Cars \cite{Krause2013data_cars},
Aircraft \cite{maji2013data_aircraft},
DTD \cite{cimpoi2014dtd}, and 
SUN397 \cite{xiao2010sun}.
They contain roughly 1k$\sim$70k training images. We follow the linear evaluation protocol \cite{kornblith2019better}. The detailed information of datasets and experimental settings are described in Appendix \ref{sec:supp:data} and \ref{section:supp:evaluation}, respectively. Table \ref{table:transfer_linear} shows the transfer learning results in the various downstream tasks. Our {\methodname} consistently improves (a) the recent unsupervised representation learning methods, SimSiam \cite{chen2020simsiam} and MoCo \cite{chen2020simsiam}, in 10 out of 11 downstream tasks; and (b) supervised pretraining in 9 out of 11 downstream tasks. These consistent improvements imply that our method encourages to learn more generalizable representations.

\textbf{Few-shot classification.}
We also evaluate the pretrained networks on various few-shot learning benchmarks: FC100 \cite{Oreshkin2018fc100}, Caltech-UCSD Birds (CUB200)~\citep{cubuk2020randaugment}, and Plant Disease \cite{mohanty2016plant}. {Note that CUB200 and Plant Disease benchmarks require low-level features such as color information of birds and leaves, respectively, to detect their fine-grained labels.} They are widely used in cross-domain few-shot settings \cite{chen2018closer_few,guo2020cdfs}. For few-shot learning, we perform logistic regression using the frozen representations $f(\mathbf{x})$ without fine-tuning. Table \ref{table:transfer_fewshot} shows the few-shot learning performance of 5-way 1-shot and 5-way 5-shot tasks. As shown in the table, our {\methodname} improves the performance of SimSiam \cite{chen2020simsiam} and MoCo \cite{he2020moco} in all cases with a large margin. For example, for plant disease detection \cite{mohanty2016plant}, we obtain up to 6.07\% accuracy gain in 5-way 1-shot tasks. These results show that our method is also effective in such transfer learning scenarios.

\textbf{Comparison with LooC.} Recently, \citet{xiao2021what} propose LooC that learns augmentation-aware representations via multiple augmentation-specific contrastive learning objectives. Table \ref{table:looc_comparison} shows head-to-head comparisons under the same evaluation setup following \citet{xiao2021what}.\footnote{Since LooC's code is currently not publicly available, we reproduced the MoCo baseline as reported in the sixth row in Table \ref{table:looc_comparison}: we obtained the same ImageNet100 result, but different ones for CUB200 and Flowers.
} As shown in the table, our {\methodname} has two advantages over LooC:
(a) {\methodname} requires the same number of augmented samples compared to the baseline unsupervised representation learning methods while LooC requires more, such that {\methodname} does not increase the computational cost; 
(b) {\methodname} can be incorporated with non-contrastive methods e.g., SimSiam \cite{chen2020simsiam}, and SimSiam with {\methodname} outperforms LooC in all cases.

\input{table_looc_comparison}

\input{fig_bird_local}
\textbf{Object localization.} We also evaluate representations in an object localization task (i.e., bounding box prediction) that requires positional information. We experiment on CUB200
\cite{van2015nabirds} and solve linear regression using representations pretrained by SimSiam \cite{chen2020simsiam} without or with our method. Table \ref{table:bird_local} reports $\ell_2$ errors of bounding box predictions and Figure \ref{figure:bird_local} shows the examples of the predictions. These results demonstrate that {\methodname} is capable of learning positional information.

\input{fig_nn}
\textbf{Retrieval.} Figure \ref{figure:nn} shows the retrieval results using pretrained models. For this experiment, we use the Flowers \cite{nilsback2008data_flowers102} and Cars \cite{Krause2013data_cars} datasets and find top-4 nearest neighbors based on the cosine similarity between representations $f(\mathbf{x})$ where $f$ is the pretrained ResNet-50 on ImageNet100. As shown in the figure, the representations learned by {\methodname} are more color-sensitive.

\subsection{Ablation study}\label{section:ablation}

\input{table_single}
\input{table_effect}

\textbf{Effect of augmentation prediction tasks.}
We first evaluate the proposed augmentation prediction tasks one by one without incorporating invariance-learning methods. More specifically, we pretrain $f_\theta$ using only $\mathcal{L}_\mathtt{aug}$ for each $\mathtt{aug}\in\{\mathtt{crop},\mathtt{flip},\mathtt{color},\mathtt{blur}\}$. Remark that training objectives are different but we use the same set of augmentations. Table \ref{table:single_task} shows the transfer learning results in various downstream tasks. 
We observe that solving horizontal flipping and Gaussian blurring prediction tasks results in worse or similar performance to a random initialized network in various downstream tasks, i.e., the augmentations do not contain task-relevant information. 
However, solving random cropping and color jittering prediction tasks significantly outperforms the random initialization in all downstream tasks. 
Furthermore,
surprisingly, the color jittering prediction task achieves competitive performance in the Flowers \cite{nilsback2008data_flowers102} dataset compared to a recent state-of-the-art method, SimSiam \cite{chen2020simsiam}. These results show that augmentation-aware information are task-relevant and learning such information could be important in downstream tasks.

Based on the above observations, we incorporate random cropping and color jittering prediction tasks into SimSiam \cite{chen2020simsiam} when pretraining. More specifically, we optimize $\mathcal{L}_\mathtt{SimSiam}+\lambda_\mathtt{crop}\mathcal{L}_\mathtt{crop}+\lambda_\mathtt{color}\mathcal{L}_\mathtt{color}$ where $\lambda_\mathtt{crop},\lambda_\mathtt{color}\in\{0,1\}$.
The transfer learning results are reported in Table \ref{table:combination}. As shown in the table, each self-supervision task improves SimSiam consistently (and often significantly) in various downstream tasks. For example, the color jittering prediction task improves SimSiam by 6.33\% and 11.89\% in Food \cite{bossard14data_food101} and Flowers \cite{nilsback2008data_flowers102} benchmarks, respectively. When incorporating both tasks simultaneously, we achieve further improvements in almost all the downstream tasks. Furthermore, as shown in Figure \ref{figure:motivation}, our {\methodname} preserves augmentation-aware information as much as possible; hence our gain is consistent regardless of the strength of color jittering augmentation. 

\input{table_new_aug}

\textbf{Different augmentations.} 
We confirm that our method can allow to use other strong augmentations: rotation, which rotates an image by $0^\circ$, $90^\circ$, $180^\circ$, $270^\circ$ degrees randomly; and solarization, which inverts each pixel value when the value is larger than a randomly sampled threshold. Based on the default augmentation setting, i.e., $\mathcal{A}_\mathtt{\methodname}=\{\mathtt{crop},\mathtt{color}\}$, we additionally apply each augmentation with a probability of $0.5$. We also evaluate the effectiveness of augmentation prediction tasks for rotation and solarization. Note that we formulate the rotation prediction as a 4-way classification task (i.e., $\omega^\mathtt{rot}_\mathtt{diff}\in\{0,1,2,3\}$) and the solarization prediction as a regression task (i.e., $\omega^\mathtt{sol}_\mathtt{diff}\in[-1,1]$). As shown in Table \ref{table:strong_aug}, we obtain consistent gains across various downstream tasks even if stronger augmentations are applied. Furthermore, in the case of rotation, we observe that our augmentation prediction task tries to prevent the performance degradation from learning invariance to rotations. For example, in CIFAR100 \citep{krizhevsky2009cifar}, the baseline loses 4.66\% accuracy (54.90\%$\rightarrow$50.24\%) when using rotations, but ours does only 0.37\% (58.65\%$\rightarrow$58.28\%). These results show that our {\methodname} is less sensitive to the choice of augmentations. We believe that this robustness would be useful in future research on representation learning with strong augmentations.

\begin{wraptable}[7]{r}{5.2cm}
\vspace{-0.15in}
\caption{Linear evaluation accuracy in augmentation-aware pretext tasks.}\label{table:pretext}
\vspace{-0.2in}
\resizebox{\linewidth}{!}{
\begin{tabular}{lcc}\\ \toprule  
Method          & Rotation & Color perm \\ \midrule
SimSiam         & 59.11    & 24.66 \\\rowcolor{Gray}
+ {\methodname} & 64.61    & 60.49 \\
\bottomrule
\end{tabular}}
\end{wraptable}
\textbf{Solving geometric and color-related pretext tasks.} To validate that our {\methodname} is capable of learning augmentation-aware information, we try to solve two pretext tasks requiring the information: 4-way rotation  ($0^\circ$, $90^\circ$, $180^\circ$, $270^\circ$) and 6-way color channel permutation (RGB, RBG, $\ldots$, BGR) classification tasks. We note that the baseline (SimSiam) and our method (SimSiam+{\methodname}) do not observe rotated or color-permuted samples in the pretraining phase.
We train a linear classifier on top of pretrained representation without finetuning for each task.
As reported in Table \ref{table:pretext}, our {\methodname} solves the pretext tasks well even without their prior knowledge in pretraining; these results validate that our method learns augmentation-aware information.

\textbf{Compatibility with other methods.} While we mainly focus on SimSiam \cite{chen2020simsiam} and MoCo \cite{he2020moco} in the previous section, our {\methodname} can be incorporated into other unsupervised learning methods, SimCLR \cite{chen2020simclr}, BYOL \cite{grill2020byol}, and SwAV \cite{caron2020swav}.
Table \ref{table:framework} shows the consistent and significant gains by {\methodname} across all methods and downstream tasks.

\input{table_frameworks}

%% file: table_imagenet100_lineval.tex
\begin{table}[t]
\centering
\caption{Linear evaluation accuracy (\%) of ResNet-50 \citep{he2016resnet} and ResNet-18 pretrained on ImageNet100 \citep{russakovsky2015imagenet,tian2019cmc} and STL10 \cite{coates2011stl10}, respectively. \textbf{Bold entries} are the best of each baseline.
}\label{table:transfer_linear}
\vspace{0.02in}
\resizebox{\textwidth}{!}{
\begin{tabular}{lccccccccccc}
\toprule
Method          & CIFAR10    & CIFAR100   & Food       & MIT67      & Pets       & Flowers    & Caltech101 & Cars       & Aircraft   & DTD        & SUN397     \\ \midrule
\multicolumn{12}{c}{\emph{ImageNet100-pretrained ResNet-50}} \\ \midrule
SimSiam         &     86.89  &     66.33  &     61.48  &     65.75  &     74.69  &     88.06  &     84.13  & {\bf48.20} &     48.63  &     65.11  &     50.60  \\ \rowcolor{Gray}
+ {\methodname} (ours) & {\bf88.80} & {\bf70.27} & {\bf65.63} & {\bf67.76} & {\bf76.34} & {\bf90.70} & {\bf85.30} &     47.52  & {\bf49.76} & {\bf67.29} & {\bf52.28} \\ \midrule
MoCo v2         &     84.60  &     61.60  &     59.37  &     61.64  &     70.08  &     82.43  &     77.25  &     33.86  & {\bf41.21} &     64.47  &     46.50  \\ \rowcolor{Gray}
+ {\methodname} (ours) & {\bf85.26} & {\bf63.90} & {\bf60.78} & {\bf63.36} & {\bf73.46} & {\bf85.70} & {\bf78.93} & {\bf37.35} &     39.47  & {\bf66.22} & {\bf48.52} \\ \midrule
Supervised      & {\bf86.16} &     62.70  &     53.89  &     52.91  &     73.50  &     76.09  & {\bf77.53} &     30.61  &     36.78  &     61.91  &     40.59  \\ \rowcolor{Gray}
+ {\methodname} (ours) &     86.06  & {\bf63.77} & {\bf55.84} & {\bf54.63} & {\bf74.81} & {\bf78.22} &     77.47  & {\bf31.26} & {\bf38.02} & {\bf62.07} & {\bf41.49} \\ \midrule
\multicolumn{12}{c}{\emph{STL10-pretrained ResNet-18}} \\ \midrule
SimSiam         &     82.35  &     54.90  &     33.99  &     39.15  &     44.90  &     59.19  &     66.33  &     16.85  &     26.06  &     42.57  &     29.05    \\ \rowcolor{Gray}
+ {\methodname} (ours) & {\bf82.76} & {\bf58.65} & {\bf41.58} & {\bf45.67} & {\bf48.42} & {\bf72.18} & {\bf72.75} & {\bf21.17} & {\bf33.17} & {\bf47.02} & {\bf34.14}   \\ \midrule
MoCo v2         &     81.18  &     53.75  &     33.69  &     39.01  &     42.34  &     61.01  &     64.15  &     16.09  &     26.63  &     41.20  &     28.50    \\ \rowcolor{Gray}
+ {\methodname} (ours) & {\bf82.45} & {\bf57.17} & {\bf36.91} & {\bf41.67} & {\bf43.80} & {\bf66.96} & {\bf66.02} & {\bf17.53} & {\bf28.02} & {\bf45.21} & {\bf30.93}   \\
\bottomrule
\end{tabular}}
\vspace{-0.1in}
\end{table}



%% file: table_imagenet100_fewshot.tex
\begin{table}[t]
\centering
\caption{
Few-shot classification accuracy (\%) with 95\% confidence intervals averaged over 2000 episodes 
on FC100 \cite{Oreshkin2018fc100}, CUB200 \cite{WahCUB_200_2011}, and Plant Disease \cite{mohanty2016plant}. $(N,K)$ denotes $N$-way $K$-shot tasks. \textbf{Bold entries} are the best of each group.}\label{table:transfer_fewshot}
\vspace{0.02in}
\scalebox{0.85}{
\begin{tabular}{lcccccc}
\toprule
& \multicolumn{2}{c}{FC100} & \multicolumn{2}{c}{CUB200} & \multicolumn{2}{c}{Plant Disease} \\
\cmidrule(lr){2-3}
\cmidrule(lr){4-5}
\cmidrule(lr){6-7}
Method           & (5, 1)              & (5, 5)              & (5, 1)              & (5, 5)              & (5, 1)              & (5, 5) \\ \midrule
\multicolumn{7}{c}{\emph{ImageNet100-pretrained ResNet-50}} \\ \midrule
SimSiam                &  \facc{36.19}{0.36} &  \facc{50.36}{0.38} &  \facc{45.56}{0.47} &  \facc{62.48}{0.48} &  \facc{75.72}{0.46} &  \facc{89.94}{0.31} \\ \rowcolor{Gray}
+ {\methodname} (ours) & \bfacc{39.37}{0.40} & \bfacc{55.27}{0.38} & \bfacc{48.08}{0.47} & \bfacc{66.27}{0.46} & \bfacc{77.93}{0.46} & \bfacc{91.52}{0.29} \\ \midrule
MoCo v2                &  \facc{31.67}{0.33} &  \facc{43.88}{0.38} &  \facc{41.67}{0.47} &  \facc{56.92}{0.47} &  \facc{65.73}{0.49} &  \facc{84.98}{0.36} \\ \rowcolor{Gray}
+ {\methodname} (ours) & \bfacc{35.02}{0.36} & \bfacc{48.77}{0.39} & \bfacc{44.17}{0.48} & \bfacc{57.35}{0.48} & \bfacc{71.80}{0.47} & \bfacc{87.81}{0.33} \\ \midrule
Supervised             &  \facc{33.15}{0.33} &  \facc{46.59}{0.37} &  \facc{46.57}{0.48} &  \facc{63.69}{0.46} &  \facc{68.95}{0.47} &  \facc{88.77}{0.30} \\ \rowcolor{Gray}
+ {\methodname} (ours) & \bfacc{34.70}{0.35} & \bfacc{48.89}{0.38} & \bfacc{47.58}{0.48} & \bfacc{65.31}{0.45} & \bfacc{70.82}{0.46} & \bfacc{89.77}{0.29} \\ \midrule
\multicolumn{7}{c}{\emph{STL10-pretrained ResNet-18}} \\ \midrule
SimSiam                &  \facc{36.72}{0.35} &  \facc{51.49}{0.36} &  \facc{37.97}{0.43} &  \facc{50.61}{0.45} &  \facc{58.13}{0.50} &  \facc{75.98}{0.40} \\ \rowcolor{Gray}
+ {\methodname} (ours) & \bfacc{40.68}{0.39} & \bfacc{56.26}{0.38} & \bfacc{41.60}{0.42} & \bfacc{56.33}{0.44} & \bfacc{62.85}{0.49} & \bfacc{81.14}{0.37} \\ \midrule
MoCo v2                &  \facc{35.69}{0.34} &  \facc{49.26}{0.36} &  \facc{37.62}{0.42} &  \facc{50.71}{0.44} &  \facc{57.87}{0.48} &  \facc{75.98}{0.40} \\ \rowcolor{Gray}
+ {\methodname} (ours) & \bfacc{39.66}{0.39} & \bfacc{55.58}{0.39} & \bfacc{38.33}{0.41} & \bfacc{51.93}{0.44} & \bfacc{60.78}{0.50} & \bfacc{78.76}{0.38} \\
\bottomrule
\end{tabular}}
\end{table}

%% file: table_looc_comparison.tex
\begin{table*}[t]
\centering
\caption{Linear evaluation accuracy (\%) under the same setup following \citet{xiao2021what}. The augmentations in the brackets of LooC \cite{xiao2021what} indicate which augmentation-aware information is learned. $N$ is the number of required augmented samples for each instance, that reflects the effective training batch size.
$^*$ indicates that the numbers are reported in \cite{xiao2021what}. 
The numbers in the brackets show the accuracy gains compared to each baseline.}\label{table:looc_comparison}
\vspace{-0.07in}
\resizebox{\textwidth}{!}{%
\begin{tabular}{lclllll}
\toprule
Method                                                & $N$ & ImageNet100 & CUB200 & Flowers (5-shot) & Flowers (10-shot)     \\ \midrule
MoCo$^*$ \cite{he2020moco}
                                                      & 2 &     81.0         &     36.7         &  \facc{67.9}{0.5}        &  \facc{77.3}{0.1}        \\
LooC$^*$ \cite{xiao2021what} (color)                  & 3 &     81.1 (+0.1)  &     40.1 (+3.4)  &  \facc{68.2}{0.6} (+0.3) &  \facc{77.6}{0.1} (+0.3) \\
LooC$^*$ \cite{xiao2021what} (rotation)               & 3 &     80.2 (-0.8)  &     38.8 (+2.1)  &  \facc{70.1}{0.4} (+2.2) &  \facc{79.3}{0.1} (+2.0) \\
LooC$^*$ \cite{xiao2021what} (color, rotation)        & 4 &     79.2 (-1.8)  &     39.6 (+2.9)  &  \facc{70.9}{0.3} (+3.0) &  \facc{80.8}{0.2} (+3.5) \\ \midrule
MoCo \cite{he2020moco}                                & 2 &     81.0         &     32.2         &  \facc{78.5}{0.3}        &  \facc{81.2}{0.3}        \\ \rowcolor{Gray}
MoCo \cite{he2020moco} + {\methodname} (ours)         & 2 &     82.4 (+1.4)  &     37.0 (+4.8)  &  \facc{81.7}{0.2} (+3.2) &  \facc{84.5}{0.2} (+3.3) \\ \midrule
SimSiam \cite{chen2020simsiam}                        & 2 &     81.6         &     38.4         &  \facc{83.6}{0.3}        &  \facc{85.9}{0.2}        \\ \rowcolor{Gray}
SimSiam \cite{chen2020simsiam} + {\methodname} (ours) & 2 & {\bf82.6} (+1.0) & {\bf45.3} (+6.9) & \bfacc{86.4}{0.2} (+2.8) & \bfacc{88.3}{0.1} (+2.4) \\
\bottomrule
\end{tabular}}
\end{table*}

%% file: fig_bird_local.tex
\begin{table}[t]
	\begin{minipage}[b][][b]{0.3\linewidth}
		\centering
		\scalebox{0.9}{
		\begin{tabular}{lr}
            \toprule
            Method & Error  \\
            \midrule
            SimSiam         & 0.00462 \\\rowcolor{Gray}
            + {\methodname} (ours) & {\bf0.00335} \\
            \midrule
            MoCo            & 0.00487 \\\rowcolor{Gray}
            + {\methodname} (ours) & {\bf0.00429} \\
            \midrule
            Supervised      & 0.00520 \\\rowcolor{Gray}
            + {\methodname} (ours) & {\bf0.00473} \\
            \bottomrule
		\end{tabular}}
		\vspace{0.05in}
		\caption{$\ell_2$ errors of bounding box predictions on CUB200.}
		\label{table:bird_local}
	\end{minipage}
	\hfill
	\begin{minipage}[b][][b]{0.66\linewidth}
		\centering
		\includegraphics[width=\linewidth]{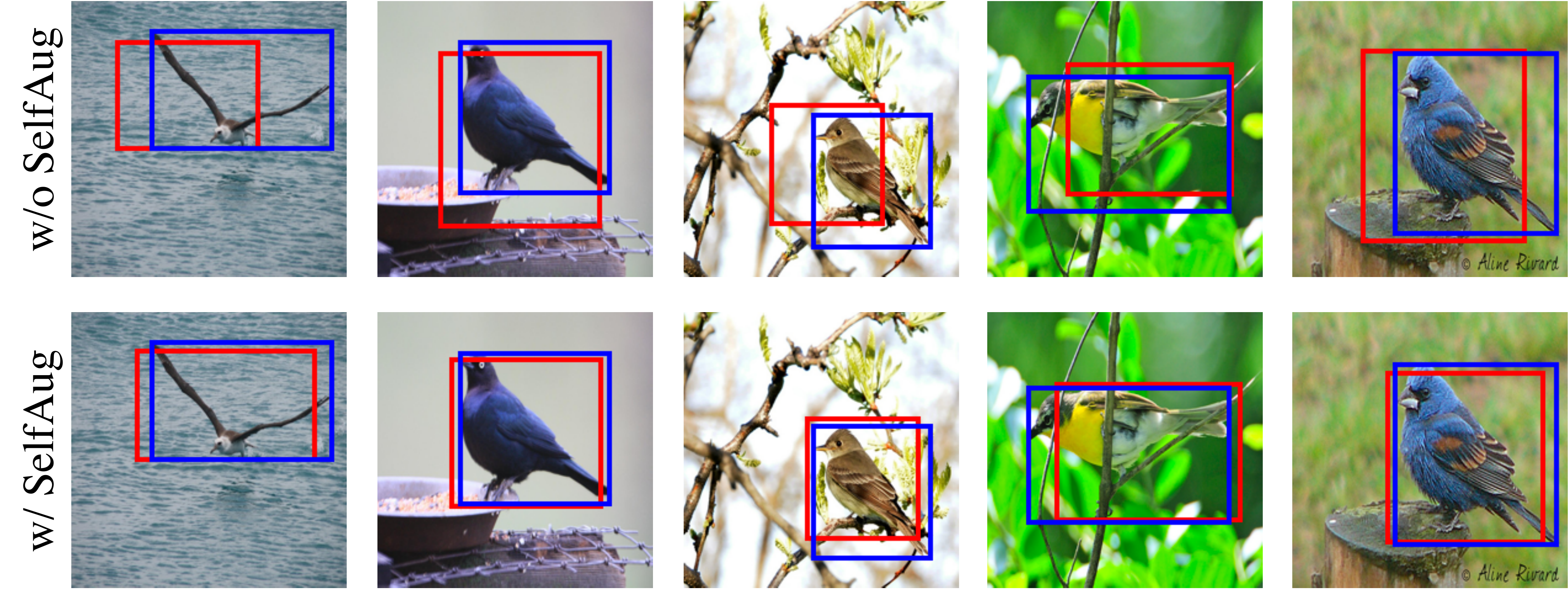}
		\captionof{figure}{Examples of bounding box predictions on CUB200. Blue and red boxes are ground-truth and model prediction, respectively.}
		\label{figure:bird_local}
	\end{minipage}
	\vspace{-0.15in}
\end{table}

%% file: fig_nn.tex
\begin{figure}[t]
\centering
\begin{subfigure}[b]{0.49\textwidth}
\centering
\includegraphics[width=0.95\textwidth]{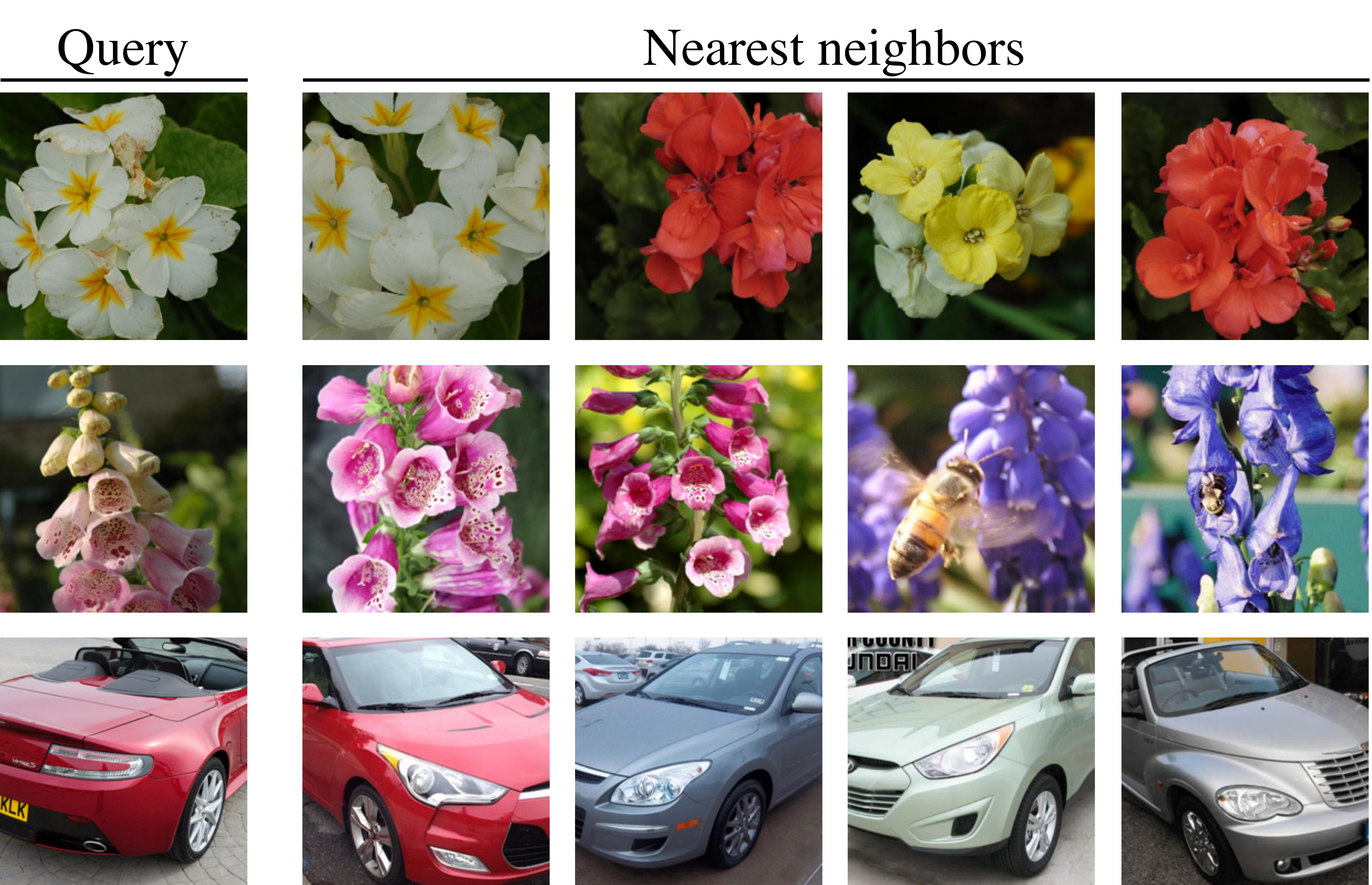}
\caption{SimSiam}\label{figure:nn:baseline}
\end{subfigure}
\hfill
\begin{subfigure}[b]{0.49\textwidth}
\centering
\includegraphics[width=0.95\textwidth]{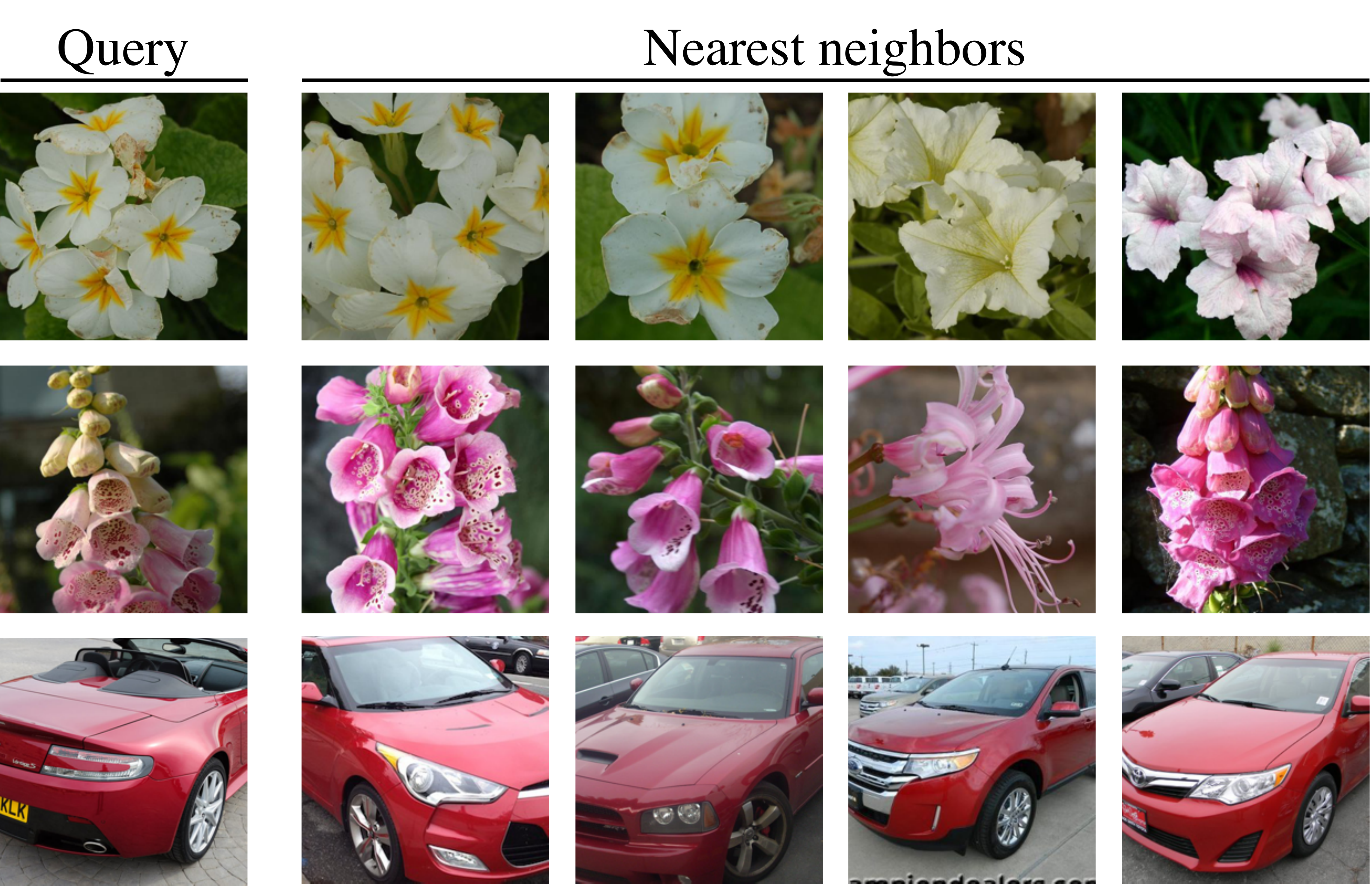}
\caption{SimSiam + {\methodname} (ours)}\label{figure:nn:ours}
\end{subfigure}
\vspace{-0.05in}
\caption{Top-4 nearest neighbors based on the cosine similarity using representations $f(\mathbf{x})$ learned by (a) SimSiam \cite{chen2020simsiam} or (b) SimSiam with {\methodname} (ours).}
\label{figure:nn}
\vspace{-0.1in}
\end{figure}

%% file: table_single.tex
\begin{table}[t]
\centering
\caption{Linear evaluation accuracy (\%) of ResNet-18 \citep{he2016resnet} pretrained by each augmentation prediction task without other methods such as SimSiam \cite{chen2020simsiam}.
We report SimSiam \cite{chen2020simsiam} results as reference.
\textbf{Bold entries} are larger than the random initialization.}\label{table:single_task}
\vspace{0.05in}
\scalebox{0.9}{%
\begin{tabular}{lccccccc}
\toprule
Pretraining objective           & STL10       & CIFAR10     & CIFAR100    & Food        & MIT67       & Pets        & Flowers     \\ \midrule
Random Init                     &      42.72  &      47.45  &      23.73  &      11.54  &      12.29  &      12.94  &      26.06  \\ \midrule
$\mathcal{L}_\mathtt{crop}$     & {\bf 68.28} & {\bf 70.78} & {\bf 43.44} & {\bf 22.26} & {\bf 26.17} & {\bf 27.68} & {\bf 38.21} \\
$\mathcal{L}_\mathtt{flip}$     & {\bf 46.45} & {\bf 53.80} & {\bf 24.89} &       9.69  &      11.99  &      10.71  &      13.04  \\
$\mathcal{L}_\mathtt{color}$    & {\bf 61.14} & {\bf 63.39} & {\bf 40.38} & {\bf 28.02} & {\bf 25.35} & {\bf 24.49} & {\bf 54.42} \\
$\mathcal{L}_\mathtt{blur}$     & {\bf 48.26} &      46.60  &      20.44  &       8.73  &      11.87  & {\bf 13.07} &      17.20  \\ \midrule
SimSiam \citep{chen2020simsiam} & {\bf 85.19} & {\bf 82.35} & {\bf 54.90} & {\bf 33.99} & {\bf 39.15} & {\bf 44.90} & {\bf 59.19} \\
\bottomrule
\end{tabular}}
\vspace{-0.05in}
\end{table}

%% file: table_effect.tex
\begin{table}[t]
\centering
\caption{Linear evaluation accuracy (\%) of ResNet-18 \citep{he2016resnet} pretrained by SimSiam \citep{chen2020simsiam} with various combinations of our augmentation prediction tasks. \textbf{Bold entries} are the best of each task.}\label{table:combination}
\vspace{0.02in}
\scalebox{0.9}{
\begin{tabular}{lccccccc}
\toprule
$\mathcal{A}_\mathtt{\methodname}$ & STL10          & CIFAR10        & CIFAR100       & Food           & MIT67          & Pets           & Flowers \\ \midrule
$\emptyset$                        & 85.19          & 82.35          & 54.90          & 33.99          &         39.15  & 44.90          & 59.19 \\
$\{\mathtt{crop}\}$                & \textbf{85.98} &         82.82  & 55.78          & 35.68          &         43.21  & 47.10          & 62.05 \\
$\{\mathtt{color}\}$               &         85.55  & \textbf{82.90} &         58.11  & 40.32          &         43.56  & 47.85          & 71.08 \\
$\{\mathtt{crop},\mathtt{color}\}$ &         85.70  &         82.76  & \textbf{58.65} & \textbf{41.58} & \textbf{45.67} & \textbf{48.42} & \textbf{72.18} \\
\bottomrule
\end{tabular}}
\vspace{-0.05in}
\end{table}

%% file: table_new_aug.tex
\begin{table}[t]
\vspace{-0.05in}
\centering
\caption{
Transfer learning accuracy (\%) of ResNet-18 \citep{he2016resnet} pretrained by SimSiam \citep{chen2020simsiam} with or without our {\methodname} using strong augmentations. $\mathtt{C}$, $\mathtt{J}$, $\mathtt{R}$ and $\mathtt{S}$ denote cropping, color jittering, rotation and solarization prediction tasks, respectively. \textbf{Bold entries} are the best of each augmentation.}\label{table:strong_aug}
\vspace{0.02in}
\scalebox{0.9}{
\begin{tabular}{clccccccc}
\toprule
Strong Aug. & $\mathcal{A}_\mathtt{\methodname}$        & STL10      & CIFAR10    & CIFAR100   & Food    & MIT67      & Pets       & Flowers \\ \midrule
\multirow{2}{*}{None}                                                     
& $\emptyset$                        &     85.19  &     82.35  &     54.90  &     33.99  &     39.15  &     44.90  &     59.19 \\
& $\{\mathtt{C},\mathtt{J}\}$            & {\bf85.70} & {\bf82.76} & {\bf58.65} & {\bf41.58} & {\bf45.67} & {\bf48.42} & {\bf72.18} \\ \midrule
\multirow{3}{*}{Rotation}                                                    
& $\emptyset$                        &     80.11  &     77.78  &     50.24  &     36.40  &     36.39  &     41.43  &     61.77  \\
& $\{\mathtt{C},\mathtt{J}\}$            &     81.85  &     79.93  &     57.27  &     43.04  &     41.32  & {\bf47.30} &     72.52  \\
& $\{\mathtt{C},\mathtt{J},\mathtt{R}\}$ & {\bf82.67} & {\bf80.71} & {\bf58.28} & {\bf43.28} & {\bf44.48} &     46.65  & {\bf72.94} \\ \midrule
\multirow{3}{*}{Solarization}                                                
& $\emptyset$                        & {\bf86.32} &     81.08  &     52.50  &     32.59  &     41.29  &     44.76  &     58.79 \\
& $\{\mathtt{C},\mathtt{J}\}$            &     86.03  & {\bf82.64} &     57.94  & {\bf40.29} & {\bf46.67} &     48.81  & {\bf71.43} \\
& $\{\mathtt{C},\mathtt{J},\mathtt{S}\}$ &     85.91  &     82.63  & {\bf58.18} &     40.17  &     45.57  & {\bf49.02} & {\bf71.43} \\
\bottomrule
\end{tabular}}
\vspace{-0.1in}
\end{table}

%% file: table_frameworks.tex
\begin{table}[t]
\centering
\caption{Transfer learning accuracy (\%) of various unsupervised learning frameworks with and without our {\methodname} framework.
\textbf{Bold entries} indicates the best for each baseline method.
}\label{table:framework}
\vspace{0.02in}
\scalebox{0.9}{
\begin{tabular}{ccccccccc}
\toprule
Method& {\methodname} (ours) & STL10             & CIFAR10    & CIFAR100   & Food    & MIT67      & Pets        & Flowers \\ \midrule
&                               &     84.87         &     78.93  &     48.94  &     31.97  &     36.82  &      43.18  &     56.20 \\ \rowcolor{Gray}
\multirow{-2}{*}{SimCLR \cite{chen2020simclr}\cellcolor{white}}                            
& \checkmark                    & {\bf84.99}        & {\bf80.92} & {\bf53.64} & {\bf36.21} & {\bf40.62} &  {\bf46.51} & {\bf64.31} \\ \midrule
&                               &     86.73         &     82.66  &     55.94  &     37.30  &     42.78  &      50.21  &     66.89 \\ \rowcolor{Gray}
\multirow{-2}{*}{BYOL \citep{grill2020byol}\cellcolor{white}}                                                       
& \checkmark                    & {\bf86.79}        & {\bf83.60} & {\bf59.66} & {\bf42.89} & {\bf46.17} &  {\bf52.45} & {\bf74.07} \\ \midrule
&                               &     82.21         &     81.60  &     52.00  &     29.78  &     36.69  &      37.68  &     53.01  \\ \rowcolor{Gray} 
\multirow{-2}{*}{SwAV \cite{caron2020swav}\cellcolor{white}}                            
& \checkmark                    & {\bf82.57}        & {\bf82.00} & {\bf55.10} & {\bf33.16} & {\bf39.13} &  {\bf40.74} & {\bf61.69} \\
\bottomrule
\end{tabular}}
\vspace{-0.1in}
\end{table}

%% file: 5_related.tex
\section{Related work}\label{section:related}

\textbf{Self-supervised pretext tasks.} For visual representation learning without labels,
various pretext tasks have been proposed in literature
\citep{doersch2015patch_location,noroozi2016jigsaw,zhang2016colorful,gidaris2018rotation,larsson2016color,qi2019avt,zhang2019aet,zhang2017splitbrain} by constructing self-supervision from an image. 
For example, \citet{doersch2015patch_location,noroozi2016jigsaw} split the original image $\mathbf{x}$ into $3\times3$ patches and then learn visual representations by predicting relations between the patch locations. Instead, \citet{zhang2016colorful,larsson2016color} construct color prediction tasks by converting colorful images to gray ones. \citet{zhang2017splitbrain} propose a similar task requiring to predict one subset of channels (e.g., depth) from  another (e.g., RGB values). Meanwhile, \citet{gidaris2018rotation,qi2019avt,zhang2019aet} show that solving affine transformation (e.g., rotation) prediction tasks can learn high-level representations. These approaches often require specific preprocessing procedures (e.g., $3\times3$ patches \citep{doersch2015patch_location,noroozi2016jigsaw}, or specific affine transformations \citep{qi2019avt,zhang2019aet}). In contrast, our {\methodname} can be working with common augmentations such as random cropping and color jittering. This advantage allows us to incorporate {\methodname} to the recent state-of-the-art frameworks like SimCLR \citep{chen2020simclr} while not increasing the computational cost. Furthermore, we emphasize that our contribution is not only to construct {\methodname}, but also finding on the importance of learning augmentation-aware representations together with the existing augmentation-invariant approaches.

\textbf{Augmentations for unsupervised representation learning.} \citet{chen2020simclr,tian2020good_view} found that the choice of augmentations plays a critical role in contrastive learning. Based on this finding, many unsupervised learning methods \citep{he2020moco,chen2020simclr,caron2020swav,grill2020byol,chen2020simsiam} have used similar augmentations (e.g., cropping and color jittering) and then achieved outstanding performance in ImageNet \cite{russakovsky2015imagenet}.
\citet{tian2020good_view} discuss a similar observation to ours that the optimal choice of augmentations is task-dependent, but they focus on finding the optimal choice for a specific downstream task. However, in the pretraining phase, prior knowledge of downstream tasks could be not available. In this case, we need to preserve augmentation-aware information in representations as much as possible for unknown downstream tasks.
Recently, \citet{xiao2021what} propose a contrastive method for learning augmentation-aware representations.
This method requires additional augmented samples for each augmentation; hence, the training cost increases with respect to the number of augmentations. Furthermore, the method is specialized to contrastive learning only, and
is not attractive to be used for non-contrastive methods like BYOL \cite{grill2020byol}. In contrast, our {\methodname} does not suffer from these issues as shown in Table \ref{table:looc_comparison} and \ref{table:framework}.

%% file: 0_appendix.tex
\section{Trade-off between augmentation invariance and awareness}\label{appendix:tradeoff}

We first emphasize that we want $f(\mathbf{x})$ to learn both augmentation-invariant and augmentation-aware information (or features) in the input $\mathbf{x}$. To this end, we train $g$ and $\phi$ to extract each information from $f(\mathbf{x})$, respectively; in other words, we want the functions $g(f(t_1(\mathbf{x})))$ and $\phi(f(t_1(\mathbf{x})), f(t_2(\mathbf{x})))$ to be invariant and variant with respect to the augmentation $t_1$ (and $t_2$), respectively. Here, if the shared network 
$f$ has a limited capacity (e.g., few parameters or dimension), the two training objectives (for $g$ and $\phi$) may interfere with each other, i.e., $f(\mathbf{x})$ might become less invariant (or contain less augmentation-invariant information). However, our choice $f$ of deep neural networks (DNNs) in our experiments does not suffer from the issue (i.e., DNNs are highly expressive), so our goal is achievable with a negligible loss of augmentation-invariant information.
To support this, we compute the cosine similarity between representations from augmented and original samples, i.e., $\mathtt{CS}=\mathbb{E}_{\mathbf{x}\sim \mathcal{D},t\sim \mathcal{T}}[\mathtt{sim}(g\circ f(t(\mathbf{x})), g\circ f(\mathbf{x}))]$. Note that this metric becomes higher as the representation $g(f(\mathbf{x}))$ is more invariant to the augmentation $t\sim\mathcal{T}$. We here use STL10-pretrained models. Table \ref{table:invariance} shows that {\methodname} does not significantly change the cosine similarity $\mathtt{CS}$; in other words, AugSelf is not harmful to the augmentation-invariant objective.

\begin{table}[ht]
\centering
\caption{The invariance metric, $\mathtt{CS}=\mathbb{E}_{\mathbf{x}\sim \mathcal{D},t\sim \mathcal{T}}[\mathtt{sim}(g\circ f(t(\mathbf{x})), g\circ f(\mathbf{x}))]$, with 95\% confidence intervals over 80k random samples in the STL10 test split \cite{coates2011stl10}. Higher values mean $f(\mathbf{x})$ is more invariant to the augmentations $\mathcal{T}$.}
\label{table:invariance}
\vspace{0.05in}
\scalebox{0.9}{%
\begin{tabular}{ccccc}
\toprule
{\methodname} (ours) & SimSiam \cite{chen2020simsiam} & BYOL \cite{grill2020byol} & SimCLR \cite{chen2020simclr} & MoCo \cite{he2020moco} \\ \midrule
                   & \facc{0.9263}{0.0005} & \facc{0.9555}{0.0004} & \facc{0.9378}{0.0006} & \facc{0.9274}{0.0006} \\\rowcolor{Gray}
\checkmark         & \facc{0.9250}{0.0006} & \facc{0.9453}{0.0004} & \facc{0.9385}{0.0005} & \facc{0.9280}{0.0006} \\
\bottomrule
\end{tabular}}
\end{table}

\section{Hyperparameter sensitivity analysis}\label{appendix:lambda}

We simply use the same value of $\lambda$, e.g., $\lambda=1$ for STL10 experiments, across different augmentations and different downstream tasks. One can find a better hyperparameter by tuning it on each augmentation and each downstream task, but we do not make much effort to tune it as our method is not too sensitive to hyperparameters. We here provide the sensitivity analysis to the hyperparameter $\lambda$ with varying $\lambda\in\{0.5,1.0,2.0\}$ under the STL10 pretraining setup. Table \ref{table:tradeoff} shows that the overall transfer learning performance is not too sensitive to $\lambda$ and {\methodname} clearly improves the performance in all the cases over the baseline (i.e., $\lambda=0$).

\input{table_tradeoff}

\newpage
\section{Fine-tuning experiments}

We fine-tune ImageNet100-pretrained models using a fine-tuning strategy, L2SP \cite{xuhong2018l2sp}.\footnote{L2SP \cite{xuhong2018l2sp} use $\Omega(\theta)=\frac{\alpha}{2}\lVert\theta-\theta^0\rVert_2^2+\frac{\beta}{2}\lVert\theta_\mathtt{cls}\rVert_2^2$ as the regularization term where $\theta^0$ is the vector of pretrained parameters (i.e., an initial point) and $\theta_\mathtt{cls}$ is the vector of classifier parameters.} Following \citet{xuhong2018l2sp}, we use the same set fo hyperparameters, i.e., $\beta=0.01$, $\alpha\in\{0.001,0.01,0.1,1\}$, and $lr\in\{0.005,0.01,0.02\}$. We evaluate the fine-tuning accuracy on five benchmarks, MIT67 \cite{quattoni2009mit67}, CUB200 \cite{WahCUB_200_2011}, Food \cite{bossard14data_food101}, Stanford Dogs \cite{Khosla20211dogs}, and Caltech256 \citep{griffin2007caltech}. Table \ref{table:finetuning} shows the effectiveness of our {\methodname} in the fine-tuning scenarios.

\begin{table}[ht]
\centering
\caption{Fine-tuning accuracy (\%) averaged over 5 trials with 95\% confidence intervals.}
\label{table:finetuning}
\vspace{0.02in}
\scalebox{0.9}{%
\begin{tabular}{lcccccc}
\toprule
Method                 & MIT67 & CUB200 & Food & Standford Dogs & Caltech256 \\ \midrule
SimSiam                &  \facc{67.69}{0.48} &  \facc{67.59}{0.41} &  \facc{66.66}{0.36} &  \facc{69.34}{0.29} &  \facc{52.37}{0.45} \\\rowcolor{Gray}
+ {\methodname} (ours) & \bfacc{69.31}{0.56} & \bfacc{70.22}{0.66} & \bfacc{70.19}{0.18} & \bfacc{70.26}{0.20} & \bfacc{54.66}{0.07} \\
\bottomrule
\end{tabular}}
\end{table}

\section{Robustness under perturbations}

We also evaluate the robustness of the learned representations by our method. To this end, we use two types of robustness metrics: (1) adversarial robustness using the single-step fast gradient sign method (FGSM) \cite{goodfellow2014explaining} and (2) robustness to common corruptions, especially weather (fog, frost, and snow) corruptions, proposed by \cite{hendrycks2018benchmarking}. We here use supervised models trained on ImageNet100 for generating adversarial samples. Table \ref{table:robustness} shows the classification accuracy on ImageNet100 under the two types of perturbations.

\begin{table}[ht]
\centering
\caption{Classification accuracy (\%) under various perturbations.}
\label{table:robustness}
\vspace{0.02in}
\scalebox{0.9}{%
\begin{tabular}{lcccccccc}
\toprule
                       &       & \multicolumn{3}{c}{FGSM} & \multicolumn{3}{c}{Weather corruption} \\ \cmidrule(lr){3-5}\cmidrule(lr){6-8}
Method                 & Clean & $\epsilon=1/255$ & $\epsilon=2/255$ & $\epsilon=4/255$ & Fog & Frost & Snow \\ \midrule
SimSiam                & 85.60 & 32.48 & 22.80 & 17.70 & 57.53 & 53.96 & 43.85 \\\rowcolor{Gray}
+ {\methodname} (ours) & 85.40 & 32.90 & 21.60 & 16.64 & 57.42 & 53.97 & 45.65 \\
\bottomrule
\end{tabular}}
\end{table}

We observe that our {\methodname} does not significantly affect the robustness of learned representations. This result is somewhat interesting because the representations learned with {\methodname} are more sensitive to diverse information than those without {\methodname}. Since improving the adversarial robustness of self-supervised learning is an ongoing topic \cite{kim2020adversarial,ho2020contrastive}, we believe that incorporating the idea with our framework would be an interesting research direction.

\section{Datasets}\label{sec:supp:data}

Table \ref{table:dataset_info} summarizes detailed descriptions of (a) pretraining datasets, (c) linear evaluation benchmarks, and (c) few-shot learning benchmarks. For linear evaluation benchmarks, we randomly choose validation samples in the training split for each dataset when the validation split is not officially provided. For few-shot benchmarks, we use the meta-test split for FC100 \citep{Oreshkin2018fc100}, and whole datasets for CUB200 \citep{WahCUB_200_2011} and Plant Disease \citep{mohanty2016plant}. The evaluation details are described in Section \ref{section:supp:evaluation}.

\input{table_dataset_info}

\section{Pretraining setup}\label{sec:supp:pretrain}

\subsection{ImageNet100 pretraining}\label{section:supp:pretrain_imagenet100}

We pretrain the standard ResNet-50 \citep{he2016resnet} architecture in the ImageNet100\footnote{ImageNet100 is a 100-category subset of ImageNet \cite{russakovsky2015imagenet}. We use the same split following \citet{tian2019cmc}.} \citep{russakovsky2015imagenet,tian2019cmc} dataset for $500$ training epochs using SimSiam \citep{chen2020simsiam} and MoCo \citep{he2020moco} methods. We use a batch size of $256$ and a cosine learning rate schedule without restarts \citep{loshchilov2016sgdr}. Note that the pretraining setups are the same as they officially used for ImageNet pretraining described in \citep{chen2020simsiam,he2020moco,chen2020moco_v2}. In multi-GPU experiments, we use the synchronized batch normalization following \citet{chen2020simsiam}. When incorporating our {\methodname} into the methods, we use $\lambda=0.5$ and $\mathcal{A}_\mathtt{\methodname}=\{\mathtt{crop},\mathtt{color}\}$. Note that SimSiam \citep{chen2020simsiam} and MoCo \citep{he2020moco} requires 32 and 29 hours on a single RTX3090 4-GPU machine, respectively.

\textbf{SimSiam} \citep{chen2020simsiam}. We use a learning rate $0.05$ and a weight decay of $0.0001$. We use a 3-layer projection MLP head $g(\cdot)$ with a hidden dimension of $2048$ and an output dimension of $2048$. We use a batch normalization \citep{ioffe2015batch_norm} at the last layer in the projection MLP. We use a 2-layer prediction MLP head $h(\cdot)$ with a hidden dimension of $512$ and no batch normalization at the last layer in the prediction MLP. When optimizing the prediction MLP, we use a constant learning rate schedule following \citet{chen2020simsiam}.

\textbf{MoCo} \citep{he2020moco}. We use a learning rate $0.03$ and a weight decay of $0.0001$. Following an advanced version of MoCo \citep[MoCo v2]{chen2020moco_v2}, we use a 2-layer projection MLP head $g(\cdot)$ with a hidden dimension of 2048 and an output dimension of $128$. We use a batch normalization \cite{ioffe2015batch_norm} at only the hidden layer. We also use a temperature scaling parameter of $0.2$, an exponential moving average parameter of $0.999$, and a queue size of $65536$.

\subsection{STL10 pretraining}\label{section:supp:pretrain_stl10}

We pretrain the standard ResNet-18 \citep{he2016resnet} architecture in the STL10 \citep{coates2011stl10} dataset. For all methods, we use the same optimization scheme: stochastic gradient descent (SGD) with a learning rate of $0.03$, a batch size of $256$, a weight decay of $0.0005$, a momentum of $0.9$. The learning rate follows a cosine decay schedule without restarts \citep{loshchilov2016sgdr}. When incorporating our {\methodname} into the methods, we use $\lambda=1.0$ and $\mathcal{A}_\mathtt{\methodname}=\{\mathtt{crop},\mathtt{color}\}$, unless otherwise stated. We now describe method-specific hyperparameters one by one in the following.

\textbf{SimCLR} \citep{chen2020simclr}. We use a 2-layer projection MLP head $g(\cdot)$ with a hidden dimension of $512$ and an output dimension of $128$. We do not use a batch normalization \citep{ioffe2015batch_norm} at the last layer in the MLP. We use a temperature scaling parameter of $0.2$ in contrastive learning.

\textbf{MoCo} \citep{he2020moco}. We use an advanced version of MoCo \citep[MoCo v2]{chen2020moco_v2} with the same projection MLP architecture as SimCLR used. Other hyperparameters are the same as the ImageNet100 setup described in Section~\ref{section:supp:pretrain_imagenet100}.

\textbf{BYOL} \citep{grill2020byol}. Following \citet{grill2020byol}, we use a 2-layer projection MLP head $g(\cdot)$ with a hidden dimension of $4096$ and an output dimension of $256$. We do not use a batch normalization \citep{ioffe2015batch_norm} at the last layer in the MLP. We use the same architecture for the prediction MLP head $h(\cdot)$. The exponential moving average parameter is increased starting from $0.996$ to $1.0$ with a cosine schedule following \citet{grill2020byol}.

\textbf{SimSiam} \citep{chen2020simsiam}. We use a 2-layer projection MLP with a hidden dimension of $2048$ and an output dimension of $2048$. Other hyperparameters are the same as the ImageNet100 setup described in Section~\ref{section:supp:pretrain_imagenet100}.

\textbf{SwAV} \citep{caron2020swav}.
We use a 2-layer projection MLP head $g(\cdot)$ with a hidden dimension of $2048$ and an output dimension of $128$ without batch normalization \citep{ioffe2015batch_norm} at the last layer in the MLP. We use $100$ prototypes, a smoothness factor of $\epsilon=0.05$ and a temperature scaling parameter of $\tau=0.1$. We do not use the multi-resolution cropping technique and the additional queue storing previous batches for simplicity.

\subsection{Augmentations}

In this section, we describe augmentations using PyTorch \citep{paszke2019pytorch} notations in the following. Note that we use random cropping, flipping, color jittering, grayscale, and Gaussian blurring unless otherwise stated.
\vspace{-0.1in}
\begin{itemize}
\setlength\itemsep{0.02em}
\item \textbf{\texttt{RandomResizedCrop}}. The scale of cropping is randomly sampled from $[0.2, 1.0]$. The cropped images are resized to $96\times96$ and $224\times224$ for STL10 \citep{coates2011stl10} and ImageNet \citep{russakovsky2015imagenet} pretraining, respectively.
\item \textbf{\texttt{RandomHorizontalFlip}}. This operation is randomly applied with a probability of $0.5$.
\item \textbf{\texttt{ColorJitter}}. The maximum strengths of brightness, contrast, saturation and hue factors are $0.4$, $0.4$, $0.4$ and $0.1$, respectively. This operation is randomly applied with a probability of $0.8$. In Section 2, we adjust the strengths by multiplying a strength factor $s$, i.e., $s>1$ means stronger color jittering than the default configuration while $s<1$ means weaker color jittering. 
\item \textbf{\texttt{RandomGrayscale}}. This operation is randomly applied with a probability of $0.2$.
\item \textbf{\texttt{GaussianBlur}}. The standard deviation is randomly sampled from $[0.1,2.0]$. The kernel size is $9\times9$ and $23\times23$ for STL10 \citep{coates2011stl10} and ImageNet \citep{russakovsky2015imagenet} pretraining, respectively. This operation is randomly applied with a probability of $0.5$.
\item \textbf{\texttt{Rotation}}. This rotates an image by $0^\circ,90^\circ,180^\circ,270^\circ$ randomly. This operation is randomly applied with a probability of $0.5$ after the default geometric augmentations are applied.
\item \textbf{\texttt{Solarization}}. This inverts each pixel value when the value is larger than a randomly sampled threshold. Formally, for an uniformly sampled threshold $\delta\sim U(0,1)$, $$x_\mathtt{new}^{(i,j)}\leftarrow\mathbbm{1}[x_\mathtt{old}^{(i,j)}<\delta]x_\mathtt{old}^{(i,j)}+\mathbbm{1}[x_\mathtt{old}^{(i,j)}\ge\delta](1-x_\mathtt{old}^{(i,j)}),$$ for all pixels $(i,j)$. This operation is randomly applied with a probability of $0.5$ right after \texttt{ColorJitter} is applied.
\end{itemize}

\section{Evaluation protocol}\label{section:supp:evaluation}

\textbf{Linear evaluation benchmarks.} We follow the same linear transfer evaluation protocol \citep{chen2020simclr,grill2020byol,kornblith2019better}; we train linear classifiers upon the frozen features extracted from $224\times224$ (or $96\times96$ for STL10 pretraining) center-cropped images without data augmentation. To be specific, images are first resized to $224$ pixels along the shorter side, and then cropped by $224\times224$ at the center of the images. Then, we minimize the $\ell_2$-regularized cross-entropy objective using L-BFGS. The regularization parameter is selected from a range of 45 logarithmically spaced values from $10^{-6}$ to $10^5$ using the validation split. After selecting the best hyperparameter, we train again the linear classifier using both training and validation splits and then report the test accuracy using the model. Note that we set the maximum number of iterations in L-BFGS as $5000$ and use the previous solution as an initial point (i.e., warm start) for the next step. 

\textbf{Few-shot benchmarks.} For evaluating representations in few-shot benchmarks, we simply perform logistic regression\footnote{\label{footnote:sklearn}We use the scikit-learn \href{https://scikit-learn.org/stable/modules/generated/sklearn.linear_model.LogisticRegression.html}{\texttt{LogisticRegression}} and \href{https://scikit-learn.org/stable/modules/generated/sklearn.linear_model.LinearRegression.html}{\texttt{LinearRegression}} modules for logistic regression and linear regression, respectively.} using the frozen representations $f(\mathbf{x})$ and $N{\times}K$ support samples without fine-tuning and data augmentation in a $N$-way $K$-shot episode.

\textbf{Object localization.} For predicting bounding box information (i.e., top left coordinates and sizes of bounding boxes), we simply perform linear regression\textsuperscript{\ref{footnote:sklearn}} using the frozen representations $f(\mathbf{x})$ and all training samples in CUB200 \cite{WahCUB_200_2011} without fine-tuning and data augmentation.

%% file: table_tradeoff.tex
\begin{table}[ht]
\centering
\caption{Linear evaluation accuracy (\%) of ResNet-18 \citep{he2016resnet} pretrained by SimSiam \cite{chen2020simsiam} and our {\methodname} with varying the hyperparameter $\lambda$.}\label{table:tradeoff}
\vspace{0.05in}
\resizebox{\textwidth}{!}{%
\begin{tabular}{lccccccccc}
\toprule
Pretraining objective           & $\lambda$ & STL10       & CIFAR10     & CIFAR100    & Food        & MIT67       & Pets        & Flowers    & Avg \\ \midrule
$\mathcal{L}_\mathtt{SimSiam}$
& $0.0$ & 85.19 & 82.35 & 54.90 & 33.99 & 39.15 & 44.90 & 59.19 & 57.09 \\ \midrule
\multirow{3}{*}{$\mathcal{L}_\mathtt{SimSiam}+\lambda\mathcal{L}_\mathtt{crop}$}
& $0.5$ & 85.50 & 82.81 & 55.50 & 35.19 & 42.79 & 45.94 & 61.24 & 58.42 \\
& $1.0$ & 85.98 & 82.82 & 55.78 & 35.68 & 43.21 & 47.10 & 62.05 & 58.95 \\
& $2.0$ & 86.36 & 82.41 & 55.29 & 36.18 & 41.91 & 47.43 & 62.28 & 58.84 \\ \midrule
\multirow{3}{*}{$\mathcal{L}_\mathtt{SimSiam}+\lambda\mathcal{L}_\mathtt{color}$}
& $0.5$ & 85.66 & 83.75 & 58.58 & 39.39 & 42.61 & 47.15 & 70.05 & 61.03 \\
& $1.0$ & 85.55 & 82.90 & 58.11 & 40.32 & 43.56 & 47.85 & 71.08 & 61.34 \\
& $2.0$ & 84.79 & 82.56 & 58.89 & 40.69 & 43.41 & 46.79 & 71.93 & 61.29 \\ \midrule
\multirow{3}{*}{$\mathcal{L}_\mathtt{SimSiam}+\lambda(\mathcal{L}_\mathtt{crop}+\mathcal{L}_\mathtt{color})$}
& $0.5$ & 86.07 & 82.67 & 57.72 & 39.92 & 43.88 & 48.86 & 70.93 & 61.43 \\
& $1.0$ & 85.70 & 82.76 & 58.65 & 41.58 & 45.67 & 48.42 & 72.18 & 62.14 \\
& $2.0$ & 84.56 & 83.08 & 59.49 & 41.72 & 44.50 & 49.04 & 72.35 & 62.11 \\
\bottomrule
\end{tabular}}
\vspace{-0.05in}
\end{table}

%% file: table_dataset_info.tex
\begin{table}[ht]
\centering
\caption{Dataset information.}\label{table:dataset_info}
\vspace{0.02in}
\resizebox{\textwidth}{!}{
\begin{tabular}{clrrrrc}
\toprule
Category & Dataset & \# of classes & Training & Validation & Test & Metric \\ \midrule
\multirow{2}{*}{(a) Pretraining}
& STL10 \citep{coates2011stl10}               & 10  & 105000 & - & - & -  \\
& ImageNet100 \citep{russakovsky2015imagenet,tian2019cmc} & 1000 & 126689 & - & - & - \\ \midrule
\multirow{12}{*}{(b) Linear evaluation}
& STL10 \citep{coates2011stl10}               &  10 &  4500 &  500 &  8000 & Top-1 accuracy \\
& CIFAR10 \citep{krizhevsky2009cifar}         &  10 & 45000 & 5000 & 10000 & Top-1 accuracy \\
& CIFAR100 \citep{krizhevsky2009cifar}        & 100 & 45000 & 5000 & 10000 & Top-1 accuracy \\
& Food    \citep{bossard14data_food101}       & 101 & 68175 & 7575 & 25250 & Top-1 accuracy \\
& MIT67 \citep{quattoni2009mit67}             &  67 &  4690 &  670 &  1340 & Top-1 accuracy \\
& Pets \citep{parkhi2012pets}                 &  37 &  2940 &  740 &  3669 & Mean per-class accuracy \\
& Flowers \citep{nilsback2008data_flowers102} & 102 &  1020 & 1020 &  6149 & Mean per-class accuracy \\
& Caltech101 \citep{fei2004calteck101}        & 101 &  2525 &  505 &  5647 & Mean Per-class accuracy \\
& Cars \citep{Krause2013data_cars}            & 196 &  6494 & 1650 &  8041 & Top-1 accuracy \\
& Aircraft \citep{maji2013data_aircraft}      & 100 &  3334 & 3333 &  3333 & Mean Per-class accuracy \\
& DTD (split 1) \citep{cimpoi2014dtd}         & 47  &  1880 & 1880 &  1880 & Top-1 accuracy \\
& SUN397 (split 1) \citep{xiao2010sun}        & 397 & 15880 & 3970 & 19850 & Top-1 accuracy \\ \midrule
\multirow{3}{*}{(c) Few-shot}
& FC100 \citep{Oreshkin2018fc100}             &  20 & - & - & 12000 & Average accuracy \\
& CUB200 \citep{WahCUB_200_2011}              & 200 & - & - & 11780 & Average accuracy \\
& Plant Disease \citep{mohanty2016plant}      &  38 & - & - & 54305 & Average accuracy \\
\bottomrule
\end{tabular}}
\end{table}